\documentclass[lettersize,journal]{IEEEtran}
\usepackage{amsmath,amsfonts}
\usepackage{array}
\usepackage[caption=false,font=normalsize,labelfont=sf,textfont=sf]{subfig}
\usepackage{textcomp}
\usepackage{stfloats}
\usepackage{verbatim}
\usepackage{graphicx}
\def\BibTeX{{\rm B\kern-.05em{\sc i\kern-.025em b}\kern-.08em
    T\kern-.1667em\lower.7ex\hbox{E}\kern-.125emX}}
\usepackage{balance}

\usepackage[hyphens]{url}  
\usepackage{xcolor}
\usepackage{array}
\usepackage{mathtools}
\usepackage{amsmath}
\usepackage{multirow}
\usepackage{tabularx}
\usepackage{mathbbol}
\usepackage{amsthm}
\usepackage{tikz}
\usepackage{hyperref}

\newtheorem{theorem}{Theorem}
\newtheorem{innercustomthm}{Theorem}
\newenvironment{customthm}[1]
  {\renewcommand\theinnercustomthm{#1}\innercustomthm}
  {\endinnercustomthm}

\newtheorem{Theorem}{Theorem}[section]
\newtheorem{Remark}[Theorem]{Remark}  

 \usepackage{algorithm,algpseudocode}
\algnewcommand\algorithmicinput{\textbf{Input:}}
\algnewcommand\algorithmicoutput{\textbf{Output:}}
\algnewcommand\Input{\item[\algorithmicinput]}%
\algnewcommand\Output{\item[\algorithmicoutput]}%
\algrenewcommand\algorithmicrequire{\textbf{Input:}}
\algrenewcommand\algorithmicensure{\textbf{Output:}}

\newcolumntype{b}{>{\hsize=0.25\hsize}X}
\newcolumntype{s}{>{\hsize=0.2\hsize}X}
\newcolumntype{P}[1]{>{\centering\arraybackslash}p{#1}}

\begin{document}
\title{Revisiting Deep Generalized Canonical Correlation Analysis}
\author{Paris~A.~Karakasis,~\IEEEmembership{Graduate Student Member,~IEEE} and Nicholas~D.~Sidiropoulos,~\IEEEmembership{Fellow,~IEEE}%
\thanks{Paris~A.~Karakasis and Nicholas~D.~Sidiropoulos are with the Department of Electrical and Computer Engineering, University of Virginia, Charlottesville, VA 22904 USA (e-mail: $\left\{\text{karakasis,~nikos}\right\}$@virginia.edu).}%
\thanks{ P.~A.~Karakasis and N.~D.~Sidiropoulos were partially supported by NSF IIS-1908070 and ECCS-2118002.}
\thanks{This paper has supplementary downloadable material available at \url{https://github.com/ParisKarakasis/Revisiting-DGCCA}. The material includes software for reproducing all the presented experimental results.}
}


\maketitle

\begin{abstract}
Canonical correlation analysis (CCA) is a classic statistical method for discovering latent co-variation that underpins two or more observed random vectors. 
Several extensions and variations of CCA have been proposed that have strengthened our capabilities in terms of revealing common random factors from multiview datasets. In this work, we first revisit the most recent deterministic extensions of deep CCA and highlight the strengths and limitations of these state-of-the-art methods. Some  methods allow trivial solutions, while others can miss weak common factors. Others overload the problem by also seeking to reveal what is {\em not common} among the views -- i.e., the private components that are needed to fully reconstruct each view. The latter tends to overload the problem and its  computational and sample complexities. Aiming to improve upon these limitations, we design a novel and efficient formulation that alleviates some of the current restrictions. The main idea is to model the private components as {\em conditionally} independent given the common ones, which enables the proposed compact formulation. In addition, we also provide a sufficient condition for identifying the common random factors. Judicious experiments with synthetic and real datasets showcase the validity of our claims and the effectiveness of the proposed approach.
\end{abstract}

\begin{IEEEkeywords}
Generalized Canonical Correlation Analysis, Deep Learning, Conditional Independence
\end{IEEEkeywords}

\section{Introduction}
\label{Section_1}
\IEEEPARstart{S}{eeking} out and reasoning about similarity enables us to abstract common patterns, learn new patterns, and ultimately reason about our world. In many applications, we nowadays have rich multimodal information about an entity, topic, or concept of interest, e.g., audio and video, images and text, or different medical sensing and imaging modalities, such as EEG, MEG, and fMRI, which can be combined to provide more complete information in support of critical decision making. The different sources of information can be considered as different ``views" of an underlying phenomenon that we are interested in analyzing to accomplish several downstream tasks, such as predicting missing pieces of information or improving the quality of the observed signals. Multiview/Multimodal learning lies under the umbrella of unsupervised learning and studies how the information of multiple jointly observed views can be fused for the aforementioned problems. Under certain assumptions, the advantages of multiview techniques for several downstream tasks have also been theoretically corroborated \cite{kakade2007multi, foster2008multi, chaudhuri2009multi}.

Canonical Correlation Analysis (CCA) is a well-known statistical tool  \cite{hotelling1992relations} that can be used to extract shared information from multiple views. CCA can be considered as a generalization of Principal Component Analysis (PCA). In a scenario with two views, CCA treats the views as different random vectors and it poses the problem of finding individual linear combinations of the views that generate a common latent random vector. CCA and Generalized CCA (GCCA -- the extension to more than two views) have found many successful applications in speech processing \cite{arora2012kernel, wang2014reconstruction, wang2015unsupervised}, communications \cite{ibrahim2019cell, ibrahim2022simple}, biomedical signal processing \cite{li2009joint, correa2010canonical, katthi2021deep, karakasis2022multi}, and many other areas.

Various attempts to extend CCA to deal with nonlinear transformations have been made over the years. Kernel CCA in \cite{lai2000kernel} was one of the first attempts, while many more have followed after adopting a probabilistic point of view, as in Deep Variational CCA \cite{wang2016deep} and Nonparametric CCA \cite{michaeli2016nonparametric}, or a deterministic one as in Deep CCA (DCCA) \cite{andrew2013deep} and other works that built upon it, as in \cite{wang2015deep, benton2017deep, lyu2020nonlinear,lyu2021understanding}. Nonlinear CCA is an active research area, but one clear lesson that has emerged is that DCCA and its variants can significantly outperform the classical linear CCA in many downstream tasks. 

In this work, we first highlight what are the advantages and limitations of the previous nonlinear GCCA approaches under a deterministic setting, and then propose a novel formulation of DCCA that bypasses most of the current limitations. We provide careful motivation and experiments on synthetic and real datasets that compare the proposed approach to the prior art and showcase its promising performance.

\subsubsection*{Reproducible Research} The codes for reproducing all the presented experimental results have been submitted, for reviewing purposes, as Supplementary material in a .rar file. As for the considered datasets, they can be automatically downloaded/generated using the provided codes.

\section{Background and Related Prior Work}

\label{Section_2}

Consider a collection of $K$ random vectors $\mathbf{x}^{(k)}\in\mathbb{R}^{D_{k}}\sim\mathcal{D}_{\mathbf{x}^{(k)}}$, for $k\in\left[K\right]:=\left\{1,\ldots,K\right\}$. When these $K$ random vectors provide different indirect ``views'' of the same latent random vector $\mathbf{g}\in\mathbb{R}^{F}\sim\mathcal{D}_{\mathbf{g}}$, estimating $\mathbf{g}$ from realizations of the views is often the main problem of interest. This problem lies at the heart of (G)CCA and many prior works have considered nonlinear extensions of (G)CCA for more or less general problem instances \cite{lyu2020nonlinear, grill2020bootstrap, zbontar2021barlow, chen2021exploring, katthi2021deep}. In this section, we consider the state-of-art formulations of this problem, which can be viewed as nonlinear extensions of the most popular formulation of GCCA: the Maximum Variance (MAX-VAR) formulation.

\subsection{(Generalized) Canonical Correlation Analysis}

Assume that we observe jointly drawn realizations of a pair of random vectors $\mathbf{x}^{(1)}$ and $\mathbf{x}^{(2)}$, i.e., $K=2$. CCA \cite{hotelling1992relations} poses the problem of estimating two maximally correlated random vectors of the form $\mathbf{z}^{(k)}=\mathbf{Q}_k^T\mathbf{x}^{(k)}$. To avoid trivial solutions, CCA additionally requires that the extracted random vectors are uncorrelated. Without loss of generality, we can assume that all random vectors $\mathbf{x}^{(k)}$ are zero-mean, since the empirical mean can be subtracted as a pre-processing step. Then, the two-view CCA problem can be mathematically expressed as
\begin{equation}
\small
\begin{split}
    \max_{\left\{\mathbf{Q}_k\in\mathbb{R}^{D_k\times F}\right\}_{k=1}^2}& \mathbb{E}\left[\text{Trace}\left(\mathbf{Q}_1^T\mathbf{x}^{(1)}\mathbf{x}^{(2)^T}\mathbf{Q}_2\right)\right]\\
    \text{s.t.}\hspace{9mm}&\hspace{-6mm}\mathbb{E}\left[\mathbf{Q}_k^T\mathbf{x}^{(k)}\mathbf{x}^{(k)^T}\mathbf{Q}_k\right]=\mathbf{I}_F, \text{ for } k=1,2,
    \end{split}
    \label{LCCA}
\end{equation}
where matrices $\mathbf{Q}_k\in\mathbb{R}^{D_k\times F}$ denote the reducing operators that will transform the observed random vectors $\mathbf{x}^{(k)}$ into random vectors, of uncorrelated components, that are maximally correlated. As a result, random vectors $\mathbf{z}^{(k)}$ can be considered as different estimates of a common latent random vector, which coincide when the maximum value of the objective in (\ref{LCCA}) is attained.

The relation between solving the CCA problem and estimating the latent common random vector becomes more clear after considering the following, equivalent when $K=2$, formulation of CCA given by
\begin{equation}
\small
\begin{split}
    \min_{{\mathbf{g}}, \left\{\mathbf{Q}_k \in\mathbb{R}^{D_k\times F}\right\}_{k=1}^K}& \sum_{k=1}^K\mathbb{E}\left[\left\|\mathbf{Q}_k^T\mathbf{x}^{(k)}-\mathbf{g}\right\|^2\right]\\
    \text{s.t.}\hspace{10mm}&\hspace{2mm}\mathbb{E}\left[\mathbf{g}\mathbf{g}^T\right]=\mathbf{I}_F,
    \end{split}
\end{equation}
where $\mathbf{g}$ denotes the common latent random vector. As one can see, this formulation of CCA naturally generalizes to multiple views and is known as the MAX-VAR formulation \cite{horst1961generalized}. After considering the subspace that is spanned from the realizations of $\mathbf{g}$, CCA can be also considered as a method for the estimation of a linear subspace which is ``common'' to the $K$ sets of random variables \cite{ibrahim2019cell}.

The generalization of CCA to multiple views is known as Generalized CCA (GCCA), and several distinct formulations have been proposed over the years. More specifically, five different formulations are presented in \cite{kettenring1971canonical} with all of them boiling down to the classical CCA formulation in (\ref{LCCA}) when $K=2$ \cite{asendorf2015informative}. Nowadays, MAX-VAR has prevailed as one of the most popular formulations of GCCA, in good part because it admits a closed form solution via eigendecomposition. Moreover, GCCA identifiability has been recently considered from the common subspace point of view in \cite{SorKanSidGCCA}.  In this manuscript we focus on the MAX-VAR formulation of GCCA.

\subsection{Deep Canonical Correlation Analysis (DCCA)}

The CCA formulation is restricted to consider only linear transformations of the observed random vectors, which cannot compensate for potentially more appropriate nonlinear transformations. Andrew et al. \cite{andrew2013deep} proposed using deep neural networks (DNNs) in order to approximate nonlinear transformations and hence overcome this limitation. The formulation proposed in \cite{andrew2013deep} is equivalent to
\begin{equation}
\small
\begin{split}
    \max_{\left\{\boldsymbol{f}_k \in {\cal C}\right\}_{k=1}^2}& \mathcal{R}_2\left(\boldsymbol{f}_1,\boldsymbol{f}_2\right):=\mathbb{E}\left[\text{Trace}\left(\boldsymbol{f}_1\left(\mathbf{x}^{(1)}\right)\boldsymbol{f}_2\left(\mathbf{x}^{(2)}\right)^T\right)\right]\\  \text{s.t.}\hspace{4mm}&\hspace{6mm}\mathbb{E}\left[\boldsymbol{f}_k\left(\mathbf{x}^{(k)}\right)\boldsymbol{f}_k\left(\mathbf{x}^{(k)}\right)^T\right]=\mathbf{I}_F,\\& \hspace{6mm}\mathbb{E}\left[\boldsymbol{f}_k\left(\mathbf{x}^{(k)}\right)\right]=\mathbf{0}_F, \text{ for } k=1,2, 
\end{split}
\label{DCCA}
\raisetag{37pt}
\end{equation}
where $\boldsymbol{f}_k:\mathbb{R}^{D_k}\rightarrow\mathbb{R}^F$ and ${\cal C}$ is the class of all functions that can be generated through a given DNN parametrization. Note that a zero-mean constraint has been added, because $\boldsymbol{f}_k\left(\mathbf{x}^{(k)}\right)$ is not automatically zero-mean even if $\mathbf{x}^{(k)}$ is. Without this constraint, the orthogonality constraints by themselves are not enough to guarantee uncorrelateness.

\subsection{Deep Canonically Correlated Autoencoders (DCCAE)}

The above formulation in (3) has the disadvantage that for rich enough classes ${\cal C}$ of functions $\boldsymbol{f}_k$, trivial solutions could satisfy the considered criterion, as we discuss in the next subsection. Later, the formulation of \cite{wang2015deep} remedied this issue, although its original motivation was different. For functions $\boldsymbol{w}_k:\mathbb{R}^{F}\rightarrow\mathbb{R}^{D_k}$, let 
\begin{equation*}
\small
\begin{split}
\mathcal{L}^{(K)}\hspace{-0.5mm}\left(\boldsymbol{f}_1,\boldsymbol{w}_1,...,\boldsymbol{f}_K,\boldsymbol{w}_K\right)\hspace{-0.5mm}=\hspace{-0.5mm}\sum_{k=1}^K\hspace{-0.2mm} \mathbb{E}\hspace{-0.5mm}\left[\left\|\mathbf{x}^{(k)}\hspace{-0.5mm}-\hspace{-0.5mm}\boldsymbol{w}_k\hspace{-0.5mm}\left(\boldsymbol{f}_k\hspace{-0.5mm}\left(\hspace{-0.5mm}\mathbf{x}^{(k)}\hspace{-0.5mm}\right)\right)\hspace{-0.5mm}\right\|^2\hspace{-0.5mm}\right]\hspace{-1mm}.
\end{split}
\end{equation*}
Instead of (3), Wang et al. \cite{wang2015deep} proposed using 
\begin{equation}
\small
\begin{split}
    \max_{\left\{\boldsymbol{f}_k, \boldsymbol{w}_k \in {\cal C}\right\}_{k=1}^2}& \left(1-\lambda\right)\mathcal{R}_2\left(\boldsymbol{f}_1,\boldsymbol{f}_2\right) - \lambda\mathcal{L}^{(2)}\left(\boldsymbol{f}_1,\boldsymbol{w}_1,\boldsymbol{f}_2,\boldsymbol{w}_2\right)\\
    \text{s.t.}\hspace{7mm}&\hspace{8mm}\text{the constraints in (3)}.
    \end{split}
    \label{DCCAE}
    \raisetag{18pt}
\end{equation}

For certain distributions, it can be shown that CCA, and hence the first term of the objective in (\ref{DCCAE}), maximizes the mutual information between the projected views \cite{borga2001canonical}. On the other hand, minimizing over the reconstruction error of the each view can be seen as maximizing a lower bound on the mutual information between the corresponding view and its learned latent representation \cite{vincent2010stacked}. As a result, the DCCAE objective offers a trade-off between maximizing the mutual information between each view and its encoding, on the one hand, and maximizing the mutual information across the encodings of the two views, on the other \cite{wang2015deep}. Since the second term in the objective tends to capture the strongest, but potentially non common, latent principal components of each view, non common information could leak in the learned embeddings. This possibility could deteriorate the quality of the learned embeddings, especially in the case of weak latent common components, as we will see in our experiments.

\subsection{Deep Generalized Canonical Correlation Analysis}
The extension of the MAX-VAR formulation to the nonlinear regime, using DNNs, was proposed in \cite{benton2017deep} under the name Deep Generalized Canonical Correlation Analysis (DGCCA).
\begin{equation}
\small
\begin{split}
    \min_{{\mathbf{g}}, \left\{ \boldsymbol{f}_k \in {\cal C}\right\}_{k=1}^K}&\hspace{-3mm} \mathcal{R}^{(K)}\left(\boldsymbol{f}_1,...,\boldsymbol{f}_K,\mathbf{g}\right):=\sum_{k=1}^K\mathbb{E}\left[\left\|\boldsymbol{f}_k\left(\mathbf{x}^{(k)}\right)-\mathbf{g}\right\|^2\right]\\
    \text{s.t.}\hspace{5.5mm}&\hspace{4.5mm}\mathbb{E}\left[\mathbf{g}\mathbf{g}^T\right]=\mathbf{I}_F, \hspace{2mm}\mathbb{E}\left[\mathbf{g}\right]=\mathbf{0}_F.
    \end{split}
    \label{DGCCA}
    \raisetag{14pt}
\end{equation}
DGCCA can be seen as a direct multiview extension of DCCA and hence it inherits the disadvantages of DCCA. Since in practice we deal with finite number of samples, let us assume $M$, the constraints of (5) translate to $\mathbf{G}^T\mathbf{G}=M\cdot\mathbf{I}_F$ and $\mathbf{G}^T\mathbf{1}_M=\mathbf{0}_F$, where the rows of  $\mathbf{G}\in\mathbb{R}^{M\times F}$ correspond to the realizations of $\mathbf{g}$. One can see that the constraints can be satisfied even if $M-F-1$ samples are mapped to $\mathbf{0}_F$, while the remaining $F+1$ to the rows of any matrix $\mathbf{U}\in\mathbb{R}^{F+1\times F}$ satisfying $\mathbf{U}^T\mathbf{U}=M\cdot\mathbf{I}_F$ and $\mathbf{U}^T\mathbf{1}_{F+1}=\mathbf{0}_F$. More generally, in the case of continuous distributions $\mathcal{D}_{\mathbf{x}^{(k)}}$, for any $\mathbf{G}$ satisfying the constraints above, we can find functions $\boldsymbol{f}_k$ that map the samples of all $\mathbf{x}^{(k)}$ to the corresponding rows of $\mathbf{G}$. As a result, trivial solutions may arise leading to non informative embeddings.

\subsection{Correlation-based learning of common and private latent random vectors }
In \cite{lyu2021understanding}, Lyu et al. consider the problem of estimating disentangled representations of common and uncommon (private per view) latent random vectors, from a pair of views, by building upon DCCA. Specifically, they consider the following multiview generative model
\begin{equation}
\small
\begin{split}
    \mathbf{x}^{(k)} = \boldsymbol{v}_{k}\left(
    \begin{bmatrix}
    \mathbf{g}\\
    \mathbf{c}^{(k)}
    \end{bmatrix}
    \right), \text{ for }k=1,2,
\end{split}
\end{equation}
where $\mathbf{g}\in\mathbb{R}^{F}$ denotes the common (shared) latent random vector and $\mathbf{c}^{(k)}\in\mathbb{R}^{L_{k}}$ denote the non-common (not shared, or {\em private}) latent random vectors. Furthermore, they assume that all the random vectors together are jointly group independent, i.e. 
\begin{equation}
\small
    p\left(\mathbf{g},\mathbf{c}^{(1)},\mathbf{c}^{(2)}\right) = p\left(\mathbf{g}\right) p\left(\mathbf{c}^{(1)}\right)p\left(\mathbf{c}^{(2)}\right).
\end{equation}
Finally, functions $\boldsymbol{v}_{k}:\mathbb{R}^{F+L_k}\rightarrow\mathbb{R}^{D_k}$ are assumed to be smooth and invertible functions.

Based on the above generative model, Lyu et al. \cite{lyu2021understanding} proposed estimating common and uncommon latent factors jointly by solving the problem
\begin{equation}
\small
\begin{split}
    \min_{\substack{\boldsymbol{f}_k\in {\cal C}_f,\\ \boldsymbol{v}_k\in {\cal C}_v,\\ \mathbf{g}}}&\max_{\substack{\phi_k\in {\cal C}_{\phi},\\ \tau_1\in {\cal C}_{\tau}}} \left(1-\lambda\right)\mathcal{R}^{(2)}\left(\boldsymbol{f}_{1_S},\boldsymbol{f}_{2_S},\mathbf{g}\right)+\lambda\mathcal{L}^{(2)}\left(\boldsymbol{f}_1,\boldsymbol{v}_1,\boldsymbol{f}_2,\boldsymbol{v}_2\right)\\[-16pt]
    &\hspace{37mm}+\beta \mathcal{V}^{(2)}\left(\boldsymbol{f}_1,\phi_1, \tau_k,\boldsymbol{f}_2,\phi_2,\tau_2\right)\vspace{-4mm}\\[2pt]
\text{s.t.}\hspace{5mm}&\hspace{16mm}\mathbb{E}\left[\mathbf{g}\mathbf{g}^T\right]=\mathbf{I}_F,\hspace{2mm}\mathbb{E}\left[\mathbf{g}\right]=\mathbf{0}_F,
    \end{split}
    \label{Lyu2}
     \raisetag{14pt}
\end{equation}
where functions $\boldsymbol{f}_k\left(\mathbf{x}^{(k)}\right)$ admit the following block decomposition $\boldsymbol{f}_k\left(\mathbf{x}^{(k)}\right)=\begin{bmatrix}
\boldsymbol{f}_{k_S}\left(\mathbf{x}^{(k)}\right)^T
\boldsymbol{f}_{k_P}\left(\mathbf{x}^{(k)}\right)^T
\end{bmatrix}^T$, with subscripts $S$ and $P$ denoting the shared and private components, respectively. The second term of the objective enforces the invertibility of functions $\boldsymbol{f}_k$. At last, for functions $\phi_{k}:\mathbb{R}^F\rightarrow\mathbb{R}$ and $\tau_{k}:\mathbb{R}^{L_k}\rightarrow\mathbb{R}$, each of the terms in the sum below 
\begin{equation}
    \small
    \begin{split}
    \mathcal{V}^{(K)}&\left(\boldsymbol{f}_1,\phi_1, \tau_1,...,\boldsymbol{f}_K,\phi_K,\tau_K\right) :=\\ &\hspace{6mm}\sum_{k=1}^K\frac{\left|\mathbb{Cov}\left[\phi_k\left(\boldsymbol{f}_{k_S}\left(\mathbf{x}^{(k)}\right)\right),\tau_k\left(\boldsymbol{f}_{k_P}\left(\mathbf{x}^{(k)}\right)\right)\right]\right|}{\sqrt{\mathbb{V}\left[\phi_k\left(\boldsymbol{f}_{k_S}\left(\mathbf{x}^{(k)}\right)\right)\right]}\sqrt{\mathbb{V}\left[\tau_k\left(\boldsymbol{f}_{k_P}\left(\mathbf{x}^{(k)}\right)\right)\right]}}.
    \end{split}
\end{equation}
is used to promote group independence between the pairs of random vectors $\mathbf{g}$ and $\mathbf{c}^{(k)}$. This can be verified after noticing that the maximum correlation between several mappings of the two blocks is minimized, for each view, and recalling that when all functions of two random variables are uncorrelated, the two random variables are independent.

When the dimensions of the latent components are known, under the generative model in (6) and the assumption in (7), the authors have shown that the criterion in (8) is capable of recovering the common and the private components up to nonlinear invertible transformations. On the other hand, when the dimensions are unknown, the possibility of modelling common components as uncommon appears, as the independence promoting term does not enforce independence between random vectors $\mathbf{c}^{(k)}$. Moreover, as we show in the sequel, the assumption in (7) can be relaxed when we are interested only in estimating $\mathbf{g}$.

\section{Proposed Framework}
\label{Section_3}
In this section, we propose a multiview generative model that links the latent and the observed random vectors, $\mathbf{g}$ and $\mathbf{x}^{(k)}$. Based on this model, we consider the problem of estimating random vector $\mathbf{g}$ from realizations of $\mathbf{x}^{(k)}$ and propose a problem formulation for learning such estimators. 

\subsection{Proposed Multiview Generative Model}
For a collection of $K$ random vectors $\mathbf{x}^{(k)}\in\mathbb{R}^{D_{k}}\sim\mathcal{D}_{\mathbf{x}^{(k)}}$, for $k\in\left[K\right]$, we propose the following generative model
\begin{equation}
\small
\begin{split}
    \mathbf{x}^{(k)} = \boldsymbol{v}_{k}\left(
    \begin{bmatrix}
    \mathbf{g}\\
    \mathbf{c}^{(k)}
    \end{bmatrix}
    \right),
    \label{GenModel}
\end{split}
\end{equation}
where $\boldsymbol{v}_{k}:\mathbb{R}^{F+L_k}\rightarrow\mathbb{R}^{D_k}$ are measurable functions, $\mathbf{g}\in\mathbb{R}^{F}\sim\mathcal{D}_{\mathbf{g}}$ is an $F$-dimensional random vector, and $\mathbf{c}^{(k)}\in\mathbb{R}^{L_{k}}\sim\mathcal{D}_{\mathbf{c}^{(k)}}$ are \textit{conditionally} independent random vectors given $\mathbf{g}$, i.e. 
\begin{equation}
\small
p\left(\mathbf{g},\mathbf{c}^{(1)},\ldots,\mathbf{c}^{(K)}\right) = p\left(\mathbf{g}\right)\prod_{k=1}^K p\left(\mathbf{c}^{(k)}|\mathbf{g}\right).
\end{equation}
The generative model above implies that
\begin{equation}
\small
\begin{split}
    p\left(\mathbf{x}^{(1)},\ldots,\mathbf{x}^{(K)}\right) &= \int p\left(\mathbf{g}\right)\prod_{k=1}^K p\left(\mathbf{x}^{(k)}\big|\mathbf{g}\right)d\mathbf{g}.
\end{split}
\end{equation}
Notice that the proposed generative model is more general that the one of Lyu et al. \cite{lyu2021understanding}, since it allows dependencies among random vectors $\mathbf{c}^{(k)}$, but also between random vectors $\mathbf{c}^{(k)}$ and $\mathbf{g}$.
Next, we consider the problem of estimating $\mathbf{g}$ from complete realizations of the views.

\subsubsection*{On estimating random vector $\mathbf{g}$}

Given joint realizations of random vectors $\mathbf{x}^{(k)}$, we consider the problem of estimating $\mathbf{g}$. Several known estimators can be used for such a task that are optimal under different criteria. For example, under the Mean Squared Error (MSE) criterion, it is known that the optimal estimator is given by the conditional expectation
\begin{equation}
\small
\mathbb{E}\left[\mathbf{g}\big|\mathbf{x}^{(1)},\ldots,\mathbf{x}^{(K)}\right].
    \label{CondExp}
\end{equation}
However, two practical concerns appear with this option. One is that approximating this function of all $\mathbf{x}^{(k)}$ can be challenging in terms of complexity and scalability. Another one is that we often wish to learn $\mathbf{g}$ from only partial realizations of the views. The conditional expectation in (\ref{CondExp}) can in principle still be used in this case, but such an approach is impractical (see discussion in the Appendix). In that direction, DGCCA explores the other extreme and poses the problem of finding optimal single-view-based estimators of $\mathbf{g}$ that achieve as strong agreement as possible. Although the resulting estimators may be suboptimal, compared to the one in (13), this is a practical alternative with respect to the aforementioned concerns.

Another advantage of the DGCCA approach is that the obtained solution directly provides estimators of $\mathbf{g}$, without having to estimate the remainder of the assumed generative model. This characteristic is consistent with the main principle in statistical learning theory for solving problems using a restricted amount of information: ``\textit{When solving a given problem, try to avoid solving a more general problem as an intermediate step}" \cite{vapnik1999nature}. Estimating the full generative model is a far more general problem which may entail much higher sample complexity, and therefore worse performance when working with limited training data, as we will verify in our experiments. 

The fact we do not have access to any realizations of $\mathbf{g}$ poses some additional challenges. For example, as we mentioned earlier, there exist trivial solutions that satisfy the criterion of DGCCA in (\ref{DGCCA}). Trivial solutions often appear in unsupervised learning problem formulations \cite{hyvarinen2019nonlinear,von2021self, lyu2021understanding}. Several approaches have been considered for addressing this issue, such as adding view reconstruction regularization terms as in (\ref{DCCAE}), or considering flow-based \cite{kingma2018glow} and entropy-based \cite{von2021self} regularizations. Moreover, there also exist certain inherently unresolvable ambiguities. Specifically, it can be shown that there exist several invertible functions $\boldsymbol{\gamma}:\mathbb{R}^F\rightarrow\mathbb{R}^F$, such that random vector $\boldsymbol{\gamma}(\mathbf{g})$ is a valid alternative representation of the common latent random vector. 

\subsection{Proposed Formulation}

Our goal is to exclude trivial solutions that could emerge with DCCA and DGCCA, but also to avoid: (i) having information leakage from the uncommon latent components to the estimates of $\mathbf{g}$ that could emerge with DCCAE in (\ref{DCCAE}), (ii) the need to estimate the private components of the different views, which are often not needed for the downstream tasks after (G)CCA – especially when these represent strong noise in the individual views.

After adopting our proposed generative model, we have that all the observed random vectors $\mathbf{x}^{(k)}$ are conditionally independent given $\mathbf{g}$. As a result, given joint realizations of $\mathbf{g}$ and $\mathbf{x}^{(l)}$, for $l\neq k$, the optimal Mean Squared Error (MSE) estimator of view $\mathbf{x}^{(k)}$ can be expressed as a function of $\mathbf{g}$. This observation motivates us to consider the following formulation of multiview DGCCA
\begin{equation}
\small
\begin{split}
    \min_{\substack{\left\{\boldsymbol{f}_k\in\mathcal{C}_f\right\}_{k=1}^K,\\ \hat{\mathbf{g}},\\ \left\{\boldsymbol{w}_k\in\mathcal{C}_w\right\}_{k=1}^K}}&\hspace{-2mm}\mathcal{B}^{(K)}\left(\boldsymbol{w}_1,\ldots,\boldsymbol{w}_K,\hat{\mathbf{g}}\right) := \sum_{k=1}^K\mathbb{E}\left[\left\|\boldsymbol{w}_k\left(\hat{\mathbf{g}}\right)-\mathbf{x}^{(k)}\right\|^2\right]\\
\text{s.t.}\hspace{6.5mm}&\hspace{9mm}\mathbb{E}\left[\hat{\mathbf{g}}\hat{\mathbf{g}}^T\right]=\mathbf{I}_F,\hspace{2mm}\mathbb{E}\left[\hat{\mathbf{g}}\right]=\mathbf{0}_F,\\
&\hspace{9mm}\mathcal{R}^{(K)}\left(\boldsymbol{f}_1,...,\boldsymbol{f}_K,\hat{\mathbf{g}}\right)=0.
    \end{split}
    \label{Prop_1}
    \raisetag{29pt}
\end{equation}
The formulation in $(\ref{Prop_1})$ is designed to guard against leakage of individual components into the estimate of ${\bf g}$, and we rigorously prove this property in the following Theorem. Note that invertible nonlinear transformations of (sharing the same MSE prediction capabilities as) random vector $\mathbf{g}$ can be obtained even from the problem formulation in $(\ref{Prop_1})$ -- this is inherently unresolvable, as discussed earlier. We therefore limit ourselves to identifying an invertible nonlinear transformation of ${\bf g}$ and we provide a sufficient condition for this to happen in the following Theorem, whose proof can be found in the Appendix.
\begin{theorem}
    Under the proposed generative model in (\ref{GenModel}), consider a solution of the optimization problem in (\ref{Prop_1}). If the following additional assumptions,
    \begin{itemize}
        \item[(i)] functions $\boldsymbol{v}_k$ are also partially invertible w.r.t. $\mathbf{g}$, i.e., there exist functions $\boldsymbol{u}_k:\mathbb{R}^{D_k}\rightarrow\mathbb{R}^F$, such that $\boldsymbol{u}_k\left(\mathbf{x}^{(k)}\right)=\mathbf{g}$, for all $\mathbf{x}^{(k)}$ and $k\in\left[K\right]$,
        \item[(ii)] there exists mean dependence between at least one pair of random vectors $\mathbf{g}$ and $\mathbf{x}^{(k')}$, i.e. $$\mathbb{E}\left[\mathbf{x}^{(k')}|\mathbf{g}\right]\neq \mathbb{E}\left[\mathbf{x}^{(k')}\right] \text{ for a } k'\in\left[K\right] \text{ and all } \mathbf{g}\in\mathbb{R}^F,$$
    \end{itemize}
    hold, then the learned encodings $\boldsymbol{f}_k\left(\mathbf{x}^{(k)}\right)$ have to be non trivial functions only of $\mathbf{g}$, i.e. $\boldsymbol{f}_k\left(\boldsymbol{v}_{k}\left(
    \begin{bmatrix}
    \mathbf{g}\\
    \mathbf{c}^{(k)}
    \end{bmatrix}
    \right)\right)=\boldsymbol{\gamma}\left(\mathbf{g}\right)$, where $\boldsymbol{\gamma}:\mathbb{R}^F\rightarrow\mathbb{R}^F$. Moreover, if
    \begin{itemize}
        \item[(iii)]the conditional expectation,
$\mathbb{E}\left[ \begin{bmatrix}\mathbf{x}^{(1)^T}, \ldots, \mathbf{x}^{(K)^T}\end{bmatrix}^T \big|\mathbf{g}\right]$, is an invertible function of $\mathbf{g}$,
    \end{itemize}then function $\boldsymbol{\gamma}$ is also invertible and the latent common random vector $\mathbf{g}$ is identifiable up to invertible nonlinearities.
\end{theorem}

\begin{Remark}
$E[\mathbf{x}^{(k)}\big|\mathbf{g}]=\phi(\mathbf{g})$ is the MMSE estimate of $\mathbf{x}^{(k)}$ given random vector $\mathbf{g}$, which is a function of $\mathbf{g}$. Our assumption that  $E[\mathbf{x}^{(k)}\big|\mathbf{g}] \neq E[\mathbf{x}^{(k)}]$ simply means that this function is non-trivial, in the sense that it is not a constant (note that the right-hand side $E[\mathbf{x}^{(k)}]$ is a constant, not a random vector). This is a very mild assumption, especially considering that we only assume it for {\em one} view. 
\end{Remark}

\begin{Remark}
Assumption (ii) would not be satisfied only if every pair of random vectors $\mathbf{g}$ and $\mathbf{x}^{(k)}$ were mean-independent. However, this would require a quite careful setting of circumstances in order to happen. To see that, consider the following example of two mean-independent random variables that are actually dependent. Let $\mathbf{z}$ be the discrete two-dimensional random vector that equiprobably has the following 4 outcomes $\left\{(0,- 1),(0,+1),(-1,0),(+1,0)\right\}$, as appears in Fig. 1. It can be easily seen that it consists of quite dependent elements as the value of the one coordinate may reveal the value of the other. For example, $\mathbf{z}[2]=0$ whenever $\mathbf{z}[1]=\pm 1$. However, we have that
\begin{equation*}
\mathbb{E}\left[\mathbf{z}\left[2\right]\big|\mathbf{z}\left[1\right]=0\right]=\mathbb{E}\left[\mathbf{z}\left[2\right]\big|\mathbf{z}\left[1\right]=\pm 1\right]=\mathbb{E}\left[\mathbf{z}\left[2\right]\right]=0.
\end{equation*}
As a result, the elements of $\mathbf{z}$, although dependent, they are mean-independent. Nevertheless, this example empirically shows that mean-independence requires a very careful design of outcomes and assigned probabilities to them, in order to take place.

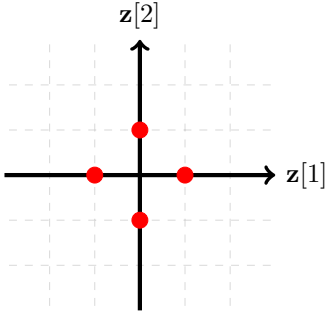
\begin{figure}
    \centering
\begin{tikzpicture}[scale=0.6]
\draw[help lines, color=gray!30, dashed] (-2.9,-2.9) grid (2.9,2.9);
\draw[->,ultra thick, black] (-3,0)--(3,0) node[right]{$\mathbf{z}[1]$};
\draw[->,ultra thick, black] (0,-3)--(0,3) node[above]{$\mathbf{z}[2]$};
\filldraw[red] (0,1) circle (5pt) node[anchor=west]{};
\filldraw[red] (0,-1) circle (5pt) node[anchor=west]{};
\filldraw[red] (-1,0) circle (5pt) node[anchor=west]{};
\filldraw[red] (1,0) circle (5pt) node[anchor=west]{};

\end{tikzpicture}
    \caption{Example of two discrete mean-independent random variables that are actually dependent.}
    \label{fig:enter-label}
\end{figure}
\end{Remark}

The appearance of random vector $\hat{\mathbf{g}}$ in the objective of (\ref{Prop_1}) but also in the constraints of (\ref{Prop_1}) complicates the design of optimization algorithms for solving it.  
For example, even if we did not have to deal with the nonlinear constraints, $\mathcal{R}^{(K)}\left(\boldsymbol{f}_1,...,\boldsymbol{f}_K,\hat{\mathbf{g}}\right)=0$, the direct approach of updating the realizations of $\hat{\mathbf{g}}$ would require methods that restrain the realizations on the Stiefel manifold, which are computationally expensive and sensitive to local minima. Next, we consider two modifications that lead to a tractable approximation of (\ref{Prop_1}). First, under the first additional assumption of Theorem 1, there exist feasible solutions, with respect to random vector $\hat{\mathbf{g}}$ and functions $\left\{\boldsymbol{f}_k\right\}_{k=1}^K$, such that the encoding of each view, $\boldsymbol{f}_k\left(\mathbf{x}^{(k)}\right)$, will be equal to $\hat{\mathbf{g}}$ in the mean square sense. As a result, the encoding of a view can be used in place of $\hat{\mathbf{g}}$ in the objective of (\ref{Prop_1}). Furthermore, we can get that
\begin{equation*}
\begin{split}
\small
\mathcal{B}^{(K)}\left(\boldsymbol{w}_1,\ldots,\boldsymbol{w}_K,\hat{\mathbf{g}}\right)&=\mathcal{Q}^{(K)}\hspace{-0.5mm}\left(\boldsymbol{f}_1,\boldsymbol{w}_1,..,\boldsymbol{f}_K,\boldsymbol{w}_K\hspace{-0.5mm}\right) \hspace{-0.5mm}\\
&:=\hspace{-0.5mm}
\frac{\displaystyle{\sum_{k=1}^K}\sum_{\substack{j=1\\j\neq k}}^K\hspace{-0.5mm}\mathbb{E}\hspace{-0.5mm}\left[\left\|\boldsymbol{w}_k\hspace{-0.5mm}\left(\boldsymbol{f}_j\hspace{-0.5mm}\left(\mathbf{x}^{(j)}\right)\right)\hspace{-0.5mm}-\hspace{-0.5mm}\mathbf{x}^{(k)}\right\|^2\hspace{-0.5mm}\right]}{(K-1)},
\end{split}
\end{equation*}
which is not a function of $\hat{\mathbf{g}}$. As for the nonlinear constraints $\mathcal{R}^{(K)}\left(\boldsymbol{f}_1,...,\boldsymbol{f}_K,\hat{\mathbf{g}}\right)=0$, we consider using a Lagrangian based approach as in the formulations of DCCAE in (\ref{DCCAE}) and of Lyu et al. in (\ref{Lyu2}).
The aforementioned modifications lead to the following tractable reformulation of (\ref{Prop_1})
\begin{equation}
\small
\begin{split}
    \min_{\substack{\left\{\boldsymbol{f}_k\in\mathcal{C}_f\right\}_{k=1}^K,\\ \hat{\mathbf{g}},\\ \left\{\boldsymbol{w}_k\in\mathcal{C}_w\right\}_{k=1}^K}}&\hspace{-4mm} \left(1-\lambda\right)\hspace{-0.5mm}\mathcal{R}^{(K)}\hspace{-0.5mm}\left(\boldsymbol{f}_1,...,\boldsymbol{f}_K,\hat{\mathbf{g}}\right)\hspace{-0.5mm}+\hspace{-0.5mm}\lambda\mathcal{Q}^{(K)}\hspace{-0.5mm}\left(\boldsymbol{f}_1,\boldsymbol{w}_1,...,\boldsymbol{f}_K,\boldsymbol{w}_K\right)\\
    \text{s.t.}\hspace{6.5mm}&\hspace{10mm}\mathbb{E}\left[\hat{\mathbf{g}}\hat{\mathbf{g}}^T\right]=\mathbf{I}_F,\hspace{2mm}\mathbb{E}\left[\hat{\mathbf{g}}\right]=\mathbf{0}_F.
    \end{split}
    \label{Prop_2}
    \raisetag{14pt}
\end{equation}
 
The intuition behind this reformulation should be clear: even if the mappings of the realizations of all the views are not equal, as long as they are close enough to $\hat{\mathbf{g}}$ they can still be used in lieu of $\hat{\mathbf{g}}$ for the reconstruction of any view. By excluding $\mathbf{x}_k$ from participating in the reconstruction of itself, we can guarantee that there will be no information leakage from the corresponding private random vector $\mathbf{c}_k$, regardless of how strong it is. In other words, by considering cross-reconstructions among all the views, we enforce capturing only the common latent factor, while the private latent components, that are potentially strong, will be ignored. As a result, the proposed formulation does not suffer from the downsides of the DCCAE method. Also, as we show next, this reformulation is convenient from an algorithmic point of view, as it enables the development of alternating optimization algorithms.

\subsection{Proposed Algorithm} 

In this section, we propose an alternating optimization algorithm for updating the estimates of the realizations of the common latent random vector $\mathbf{g}$, as well as the nonlinear mappings $\boldsymbol{f}_k$ and $\boldsymbol{w}_k$. Next, we describe the corresponding optimization subproblems and the proposed algorithm.

\subsubsection{Updating functions $\boldsymbol{f}_k$ and $\boldsymbol{w}_k$}

For fixed estimates (realizations) of random vector $\mathbf{g}$, the problem of updating functions $\boldsymbol{f}_k$ and $\boldsymbol{w}_k$ is given by
\begin{equation}
\small
\begin{split}
    \min_{\substack{\left\{\boldsymbol{f}_k\in\mathcal{C}_f\right\}_{k=1}^K,\\ \left\{\boldsymbol{w}_k\in\mathcal{C}_w\right\}_{k=1}^K}}&\hspace{-4mm} \left(1-\lambda\right)\mathcal{R}^{(K)}\left(\boldsymbol{f}_1,\ldots,\boldsymbol{f}_K,\mathbf{g}\right)\\[-20pt]
    &\hspace{24mm}+\lambda\mathcal{Q}^{(K)}\left(\boldsymbol{f}_1,\boldsymbol{w}_1,\ldots,\boldsymbol{f}_K,\boldsymbol{w}_K,\mathbf{g}\right).\\[4pt]
    \end{split}
    \label{Sub_prob_A}
    \raisetag{30pt}
\end{equation}
This is an unconstrained problem that can be tackled using stochastic optimization methods.

\subsubsection{Updating the realizations of $\hat{\mathbf{g}}$}
For fixed functions $\boldsymbol{f}_k$ and $\boldsymbol{w}_k$, the problem of updating the realizations of $\hat{\mathbf{g}}$ is given by \vspace{-4mm}
\begin{equation}
\small
\begin{split}
    \min_{{\hat{\mathbf{g}}}}& \sum_{k=1}^K\mathbb{E}\left[\left\|\boldsymbol{f}_k\left(\mathbf{x}^{(k)}\right)-\hat{\mathbf{g}}\right\|^2\right]\\
    \text{s.t.}\hspace{1mm}&\hspace{1mm}\mathbb{E}\left[\hat{\mathbf{g}}\hat{\mathbf{g}}^T\right]=\mathbf{I}_F,\hspace{2mm}\mathbb{E}\left[\hat{\mathbf{g}}\right]=\mathbf{0}_F.
    \end{split}
    \label{Sub_prob_B}
\end{equation}
Given jointly drawn realizations of random vectors $\mathbf{x}^{(k)}$, $\mathbf{x}_m^{(k)}$, the goal is to estimate the corresponding realization $\mathbf{g}_m$, for $m\in\left[M\right]$. Consider the matrices $\mathbf{G}$ and $\mathbf{Y}\in\mathbb{R}^{M\times F}$, where $\mathbf{G}(m,:) = \hat{\mathbf{g}}_m$ and
$\mathbf{Y}(m,:)=\sum_{k=1}^K \boldsymbol{f}_k\left(\mathbf{x}^{k}_m\right)$.
Then, updating the realizations of $\mathbf{g}$ boils down to the Orthogonal Procrustes problem \cite{schonemann1966generalized}
\begin{equation}
\small
\begin{split}
    \max_{\mathbf{G}\in\mathbb{R}^{M\times F}}
    \text{Trace}\left(\mathbf{G}^T\bar{\mathbf{Y}}\right),~ \text{s.t.}\hspace{2mm}\mathbf{G}^T\mathbf{G}=\mathbf{I}_F,
    \end{split}
\end{equation}
where $\bar{\mathbf{Y}}$ denotes the columnwise centered matrix $\mathbf{Y}$. Given the Singular Value Decomposition (SVD) of $\bar{\mathbf{Y}}=\mathbf{U}\boldsymbol{\Sigma}\mathbf{V}^T$, an optimal solution can be computed as $\mathbf{G}^o = \mathbf{U}\mathbf{V}^T$.

\subsubsection{Algorithm and Complexity}
The proposed algorithm is outlined in Algorithm \ref{Alg_1}. The update of $\mathbf{G}$ requires the computation of SVD with complexity $\mathcal{O}\left(MF^2\right)$. Performing a stochastic update of $\boldsymbol{f}_k$ and $\boldsymbol{w}_k$, based on a $\left|\mathcal{B}\right|$-sized batch, has complexity  $\mathcal{O}\left(\left|\mathcal{B}\right|\sum_{k=1}^Kd_k\right)$, where $d_k$ denotes the number of parameters used for approximating $\boldsymbol{f}_k$ and $\boldsymbol{w}_k$. Hence, the overall complexity of the proposed algorithm is  $\mathcal{O}\left(MF^2+\left|\mathcal{B}\right|\sum_{k=1}^Kd_k\right)$ per iteration. As for the stopping criteria, we use completing a predetermined number of iterations for the outer loop and a full epoch for the inner one.

\begin{algorithm}
\small
  \caption{Proposed Algorithm}
        \label{Alg_1}
        \begin{algorithmic}
            \Require $\left\{ \mathbf{x}^{(k)}_m\right\}_{m=1}^M,\boldsymbol{f}_k, \boldsymbol{w}_k, \forall\, k\in\left[K\right]$\vspace{1mm}
            \Ensure $\boldsymbol{f}_k^*, \boldsymbol{w}_k^* ~\forall k\in\left[K\right]$\vspace{1mm}
            \While{\textit{ stopping criterion is not met}}\vspace{1mm}
            \State - Update matrix $\mathbf{G}$ by solving (\ref{Sub_prob_B})\;\vspace{1mm}
            \While{\textit{ stopping criterion is not met}}\vspace{1mm}
                \State - Draw at random $\mathcal{B}=\left\{\mathbf{x}^{(k)}_m,\forall~k\in\left[K\right]\right\}_{m\in\mathcal{M}'\subset\left[M\right]}$\;\vspace{1mm}
                \State - Perform a stochastic gradient update of\\ \hspace{12mm}$\boldsymbol{f}_k$ and $\boldsymbol{w}_k,\forall~k\in\left[K\right]$, by restricting (\ref{Sub_prob_A}) on $\mathcal{B}$\;\vspace{1mm}
    		\EndWhile\vspace{1mm}
    		\EndWhile
        \end{algorithmic}
\end{algorithm}

\section{Experimental Results}

In this section, we present the experimental results that emerged after comparing the proposed framework to the state-of-the-art methods on three datasets; one synthetic and two real world. 
For our experiments, we have extended all the methods that are restricted to only two views by substituting the terms that coincide with the objective of DCCA to the one of DGCCA. Regarding the method proposed in \cite{lyu2021understanding}, such an extension can be achieved in a straightforward way by considering an extra set of terms in (8) for the third view. In all the experiments and for all the methods, all the MSE terms in the corresponding objectives have been normalized to reflect the relative MSE. As a result, a fair comparison of all the methods over the same value of $\lambda$ is enabled.

\subsection{Synthetic Dataset}
We begin the comparison of all the methods on a two-view synthetic dataset. For its generation, we consider the following setting. First, a sample is drawn from a categorical random variable $Z$, with alphabet $\left\{0,1,2,3\right\}$ and corresponding probabilities $0.1, 0.2, 0.3$, and $0.4$. As common random vector $\mathbf{g}$, we consider the one-hot encoding of $Z$. Regarding the private random vectors, we assume that each $\mathbf{c}^{(k)}\in\mathbb{R}^4$ follows a mixture of four zero mean Gaussian distributions with random covariance matrices. For each view, the Gaussian distribution specified by the drawn value of $Z$ is sampled to obtain the realization of the corresponding random vector $\mathbf{c}^{(k)}$. The power ratio between common and uncommon random vectors is set to $-18\,$dB. 

Based on the above procedure, we form the latent training, validation, and testing datasets, consisting of $3,000$, $1,500$, and $1,500$ samples, respectively. These samples are mapped through functions $\boldsymbol{v}_k(\cdot):\mathbf{R}^{8}\rightarrow\mathbf{R}^{64}$, for $k=1,2$, so as to create samples of the two views. For the construction of each function $\boldsymbol{v}_k$, we use a three-hidden-layer neural network with $32$ neurons per layer. The first two layers are followed by a ReLU activation function, while all the network weights are drawn i.i.d. from the standard normal distribution.

In order to approximate functions $\boldsymbol{f}_k$ and $\boldsymbol{w}_k$, we use the same architecture as above. To train them, we use the AdamW optimizer \cite{loshchilov2017decoupled} with initial learning rate equal to $0.001$ and a regularization parameter equal to $0.001$. We set the batch size equal to $100$ samples and we run the algorithm for $40$ epochs. For the method of Lyu et al. \cite{lyu2021understanding}, we set the batch size for the maximization problem equal to $300$ and the learning rate equal to $0.01$. Moreover, we approximate $\tau_k$ and $\phi_k$ using the same architecture as described above.
For the regularization parameter of the independence-promoting term, we consider two different settings. Note that random vectors $\mathbf{c}^{(k)}$ are not independent under the considered generative model. They are {\em conditionally} independent given $\mathbf{g}$, but this allows for (unconditional) dependence. This is not in agreement with the generative model of Lyu et al. \cite{lyu2021understanding}. For this reason, we consider two different values of the regularization parameter for the independence-promoting term: $0$ and $0.001$ (the independence-promoting term is ignored/considered, respectively). We employ an early stopping policy, by keeping the model that achieved the lowest objective value on the validation dataset throughout the considered number of iterations.

The embedded views ideally should convey only information related to the value of $Z$, which can be considered as label. Hence, all the pairs sharing the same value/label should have similar embeddings in the learned latent space. We compare the proposed method to all the state-of-the-art methods in terms of how well it encodes the label information. Towards this end, we follow \cite{wang2015deep} and \cite{lyu2021understanding} and consider a supervised and an unsupervised test. In more detail, we train the models of all methods for $\lambda$ taking all the values in $\left\{0.1, 0.3, 0.5, 0.7, 0.9\right\}$. For each value of $\lambda$ and method, we consider $10$ different initializations. For each learned model, we compute the averages of the embeddings of the two views for the three datasets (training/validation/testing). 

Regarding the unsupervised test, we use K-means in order to split the learned embeddings into $4$ clusters and evaluate how well the clusters agree with ground-truth labels. In the previous step, K-means is run 10 times with random initialization and the run with the best K-means objective is finally considered. We measure the clustering performance with three criteria, clustering accuracy (ACC), normalized mutual information (NMI), and adjusted Rand index (ARI) \cite{yeung2001details}. For the supervised test, we train a linear support vector machine (SVM) using the averaged embeddings of the training data. The performance of the trained SVM is evaluated on the averaged embeddings of the testing dataset and measured by the achieved classification accuracy (CLA-ACC). Finally, the final performance metrics of each method, for each value of $\lambda$, are determined by computing the averages over all $10$ corresponding learned models.

\begin{figure}[t!]
    \centering
    \includegraphics[scale = 0.5]{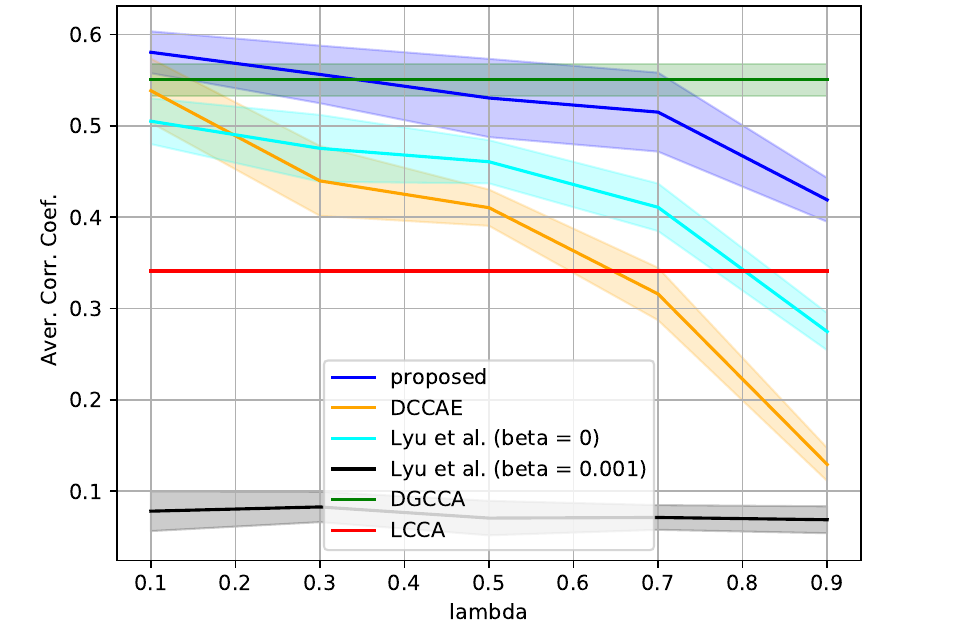}
    \caption{Average Correlation Coefficient for the synthetic dataset across different values of the $\lambda$ parameter.\\}
    \label{fig:SynthCorr}
    \centering
    \includegraphics[scale = 0.45]{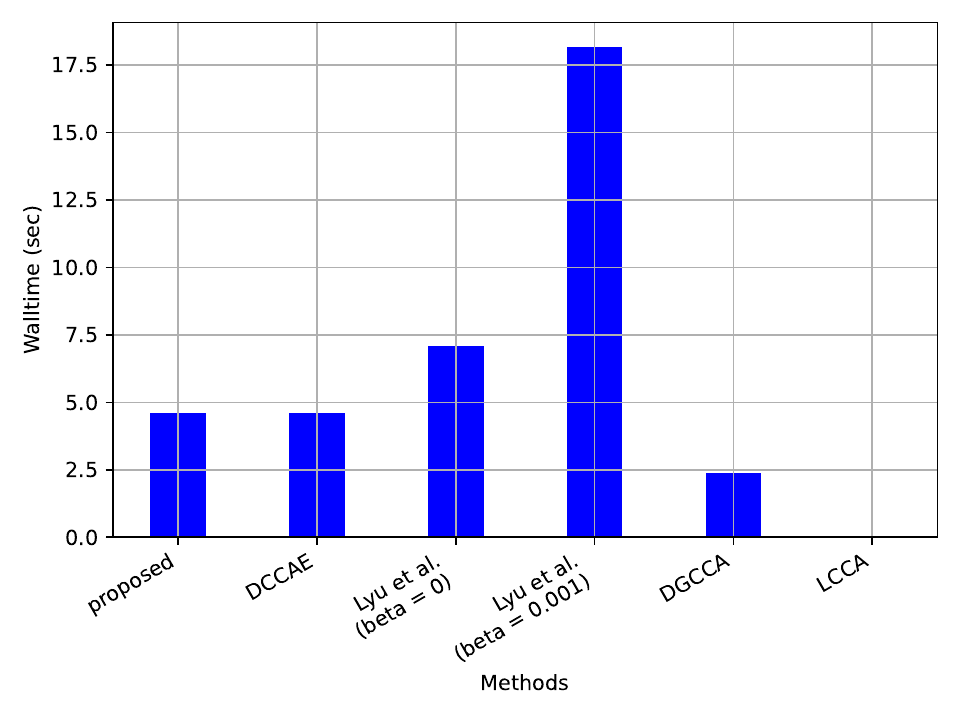}
    \caption{Average walltimes of all the considered methods for the synthetic dataset.}
    \label{fig:SynthTime}
\end{figure}

In Tables \ref{tab:ACC1}, \ref{tab:NMI1}, \ref{tab:ARI1}, and \ref{tab:CACC1}, we present the performance metrics of all methods considered versus the various values of $\lambda$.
Since CCA and DGCCA do not depend on the $\lambda$ parameter, we present the corresponding results only in the forth column ($\lambda=0.5$). We can observe that the proposed method achieves the highest scores across all the metrics, and is also the least sensitive with respect to the choice of $\lambda$. In Fig. \ref{fig:SynthCorr}, we present the average correlation coefficient that each method achieved, over the 10 learned models, for each value of $\lambda$. Regarding CCA and DGCCA, for the reason explained above, their results are depicted here as horizontal lines. We observe that the proposed method also achieves the highest correlation coefficient.

Finally, in Fig. (\ref{fig:SynthTime}) we depict the average walltimes of all the considered methods for the synthetic dataset. The linear CCA method (denoted as LCCA in the two figures) is much faster than all the other methods, but, as we discussed above, its performance is restricted by its limited capabilities in modeling nonlinear transformations. We can also observe that DGCCA is the next faster, which is expected since DGCCA does not learn functions $\boldsymbol{w}_k$ that allow the reconstruction of each view and as we discussed earlier, allows the possibility of trivial solutions.
The proposed algorithm has comparable speed with the DCCAE method, while the method of Lyu et al. \cite{lyu2021understanding} is slower than all the other methods, especially when the independence promoting term is activated where it is significantly much slower. This is expected since the algorithmic complexity analysis that we provide in the main paper holds for all the deep methods and shows that the complexity of each method is proportional to the total number of parameters that each method uses to model the corresponding encoding and decoding functions. Since the method of  \cite{lyu2021understanding} needs significantly more parameters in order to approximate the full autoencoders of the views, but also the functions for promoting independence, the resulting computational complexity is much higher. 

\begin{table}[t]
\centering
\caption{\label{tab:ACC1} Clustering accuracy (ACC) for the synthetic dataset across different values of the $\lambda$ parameter.\\ }
\resizebox{.485\textwidth}{!}{%
\begin{tabularx}{1.268\linewidth}{|P{1.49cm}|P{1.42cm}|P{1.42cm}|P{1.42cm}|P{1.42cm}|P{1.42cm}|}
\hline
  & $\lambda=0.1$ & $\lambda=0.3$ & $\lambda=0.5$ & $\lambda=0.7$ & $\lambda=0.9$\\
 \hline
\text{CCA}  & - &  -  &  0.92$\pm$0  &  - &  - \\
\hline
\text{DGCCA}  & -  & -  & 0.97$\pm$0.01  & -  & - \\
\hline
\text{DCCAE} &0.96$\pm$0.05 &0.86$\pm$0.05 & 0.84$\pm$0.05 & 0.74$\pm$0.03 & 0.48$\pm$0.06 \\
\hline
Lyu et al. (2021) $\beta=0$  &0.97$\pm$0.01 & 0.95$\pm$0.03 & 0.95$\pm$0.02 & 0.88$\pm$0.05 & 0.75$\pm$0.04 \\
\hline
Lyu et al. (2021) $\beta=0.001$ &0.40$\pm$0.06 & 0.39$\pm$0.04 & 0.40$\pm$0.03 & 0.36$\pm$0.03 & 0.38$\pm$0.03   \\
\hline
Proposed & \textbf{0.98}$\boldsymbol{\pm}$\textbf{0.01} &0.97$\pm$0.01 & 0.95$\pm$0.04 & 0.95$\pm$0.04 & 0.90$\pm$0.04 \\
\hline
\end{tabularx}%
}\vspace{5mm}
\centering
\caption{\label{tab:NMI1} Normalized mutual information (NMI) for the synthetic dataset across different values of the $\lambda$ parameter.\\ }
\resizebox{.485\textwidth}{!}{%
\begin{tabularx}{1.268\linewidth}{|P{1.49cm}|P{1.42cm}|P{1.42cm}|P{1.42cm}|P{1.42cm}|P{1.42cm}|}
\hline
  & $\lambda=0.1$ & $\lambda=0.3$ & $\lambda=0.5$ & $\lambda=0.7$ & $\lambda=0.9$\\
 \hline
CCA  & - &  -  &  0.77$\pm$0 &  - &  - \\
\hline
DGCCA  & -  & -  & 0.90$\pm$0.02  & -  & - \\
\hline
DCCAE &0.89$\pm$0.04 & 0.80$\pm$0.03 & 0.77$\pm$0.03 & 0.63$\pm$0.05 & 0.26$\pm$0.05\\
\hline
Lyu et al. (2021) $\beta=0$ &0.90$\pm$0.02 & 0.87$\pm$0.05 &0.87$\pm$0.03 &0.80$\pm$0.04 &0.61$\pm$0.05\\
\hline
Lyu et al. (2021) $\beta=0.001$ &0.10$\pm$0.06 &0.08$\pm$0.04 &0.08$\pm$0.02& 0.08$\pm$0.04 & 0.08$\pm$0.04\\
\hline
Proposed & \textbf{0.92}$\boldsymbol{\pm}$\textbf{0.03} & 0.90$\pm$0.03 & 0.88$\pm$0.04 & 0.87$\pm$0.04 & 0.78$\pm$0.02\\
\hline
\end{tabularx}%
}\vspace{5mm}
\centering
\caption{\label{tab:ARI1} Adjusted Rand Index (ARI) for the synthetic dataset across different values of the $\lambda$ parameter. \\ }
\resizebox{.485\textwidth}{!}{%
\begin{tabularx}{1.268\linewidth}{|P{1.49cm}|P{1.42cm}|P{1.42cm}|P{1.42cm}|P{1.42cm}|P{1.42cm}|}
\hline
  & $\lambda=0.1$ & $\lambda=0.3$ & $\lambda=0.5$ & $\lambda=0.7$ & $\lambda=0.9$\\
 \hline
\text{CCA}  & - &  -  &  0.84$\pm$0  &  - &  - \\
\hline
\text{DGCCA}  & -  & -  & 0.94$\pm$0.01  & -  & - \\
\hline
\text{DCCAE} &0.93$\pm$0.05 &0.81$\pm$0.06 &0.78$\pm$0.03 &0.63$\pm$0.06 &0.24$\pm$0.08\\
\hline
Lyu et al. (2021) $\beta=0$ &0.94$\pm$0.02 & 0.91$\pm$0.05 & 0.91$\pm$0.03 & 0.82$\pm$0.05 & 0.66$\pm$0.03\\
\hline
Lyu et al. (2021) $\beta=0.001$ &0.07$\pm$0.06 & 0.05$\pm$0.03 & 0.06$\pm$0.02 &0.04$\pm$0.02 &0.05$\pm$0.03\\
\hline
Proposed & \textbf{0.95}$\boldsymbol{\pm}$\textbf{0.02} & 0.94$\pm$0.02 & 0.91$\pm$0.05 & 0.91$\pm$0.05 & 0.84$\pm$0.04\\
\hline
\end{tabularx}%
}\vspace{5mm}
\centering
\caption{\label{tab:CACC1} Classification accuracy (CLA-ACC) for the synthetic dataset across different values of the $\lambda$ parameter.\\ }
\resizebox{.485\textwidth}{!}{%
\begin{tabularx}{1.268\linewidth}{|P{1.49cm}|P{1.42cm}|P{1.42cm}|P{1.42cm}|P{1.42cm}|P{1.42cm}|}
\hline
  & $\lambda=0.1$ & $\lambda=0.3$ & $\lambda=0.5$ & $\lambda=0.7$ & $\lambda=0.9$\\
 \hline
\text{MAX-VAR}  & - &  -  &  0.94$\pm$0  &  - &  - \\
\hline
\text{DGCCA}  & -  & -  & 0.98$\pm$0.01  & -  & - \\
\hline
\text{DCCAE} &0.98$\pm$0.01 &0.93$\pm$0.03 & 0.91$\pm$0.02 & 0.87$\pm$0.01 & 0.67$\pm$0.04\\
\hline
Lyu et al. (2021) $\beta=0$ &0.98$\pm$0.01 & 0.97$\pm$0.01 & 0.97$\pm$0.01 &0.94$\pm$0.02 &0.85$\pm$0.03 \\
\hline
Lyu et al. (2021) $\beta=0.001$ & 0.54$\pm$0.05 & 0.54$\pm$0.06 & 0.53$\pm$0.05 & 0.50$\pm$0.07 & 0.53$\pm$0.04\\
\hline
Proposed &\textbf{0.99}$\boldsymbol{\pm}$\textbf{0.01} &0.98$\pm$0.01 &0.97$\pm$0.03 & 0.97$\pm$0.02 & 0.93$\pm$0.02\\
\hline
\end{tabularx}%
}
\end{table}

\subsection{Acoustic-articulatory data for speech recognition}

Next, we consider a dataset from the University of Wisconsin X-Ray Micro-Beam Database (XRMB) \cite{westbury1994x}. XRMB consists of simultaneously recorded speech and articulatory measurements from 47 American English speakers. 
From those measurements, Wang et al. \cite{wang2014reconstruction, wang2015unsupervised} created a multiview dataset\footnote{The dataset can be found here: \url{https://home.ttic.edu/~klivescu/XRMB_data/full/README}}, which has been used for multiview based phonetic recognition \cite{arora2012kernel, wang2015unsupervised, wang2015deep, benton2017deep}. It consists of two views, where acoustic and articulatory features have been concatenated over a $7$-frame window around each frame, giving $273$-dimensional acoustic and $112$-dimensional articulatory features, respectively. Since each pair of samples here is tied together by two factors, the label of the spoken phoneme and the identity of the speaker, the authors of \cite{arora2014multi} and \cite{benton2017deep} considered using one-hot vector encodings of the phoneme labels $(0-39)$ as a third view, in order to find latent embeddings that convey information only related to the phoneme label. Finally, three datasets are provided for training, validation, and testing purposes, respectively, where each one of them contains samples of non-overlapping speakers. 

In our experiments, we adopt the same framework and consider the three-view setup for phonetic recognition. To limit the experiment runtime, we follow \cite{benton2017deep} and we use a subset of speakers for our experiments. Specifically, we use as training dataset the samples that correspond to speakers 1 to 5, while we use the remaining two datasets (validation and testing) intact. The underlying task of our experiments is to exploit the latent embeddings for supervised phonetic recognition, as in \cite{arora2012kernel, wang2015unsupervised, wang2015deep, benton2017deep}.
Following \cite{wang2015deep} and \cite{benton2017deep}, we consider a dimension equal to 30 for the common factor $\mathbf{g}$. For the approximation of functions $\boldsymbol{w}_k$ and $\boldsymbol{f}_k$, we use three-hidden layer neural networks with 512 neurons per layer, where the first two layers are followed by a ReLU activation function. For the training of all the models, we use the AdamW optimizer \cite{loshchilov2017decoupled} with an initial learning rate of $0.001$ and a regularization parameter equal to $0.001$. We set the batch size equal to 500 samples and we run the algorithm for 200 epochs.
For the method proposed of Lyu et al. \cite{lyu2021understanding}, we set the batch size for the maximization problem equal to $1000$ and the learning rate equal to $10^{-6}$. The dimensions of the individual components, for the first two views, are set to be equal to $60$ and $20$, after verifying that the auto-reconstruction errors are satisfactory low ($<1$\%). For the approximation of functions $\tau_k$ and $\phi_k$, we use three-hidden layer neural networks with 64 neurons per layer, where the first two layers are followed by a ReLU activation function. As the third view conveys no private information, neither private latent factors nor independence-promoting terms are considered for the third view. Regarding the regularization parameter for the independence-promoting term, we set it equal to $10^{-6}$. Finally, we adopt an early stopping policy as we did in the previous subsection. 

We train the models for all methods for regularization parameter $\lambda$ $\in$ $\left\{0.1, 0.3, 0.5, 0.7, 0.9\right\}$. For each value and method, we consider $5$ different initializations. We test each learned model on the classification task, by using the K-nearest neighbors algorithm (with K=5) on the averages of the embeddings of all three views and evaluating them on the averages of the embeddings of the first two views since the third view relies on the phoneme labels. Finally, we consider as performance metric the classification accuracy score after averaging all $5$ learned models for each method. Since the concept of correlation coefficient does not naturally extend to more than two random vectors, without having access to their joint distribution, we consider the following metric for measuring the attained correlation among the three views. For each pair of views, we compute the total correlation coefficient defined as the average of the cosines of the canonical angles between the subspaces spanned by the two views. Then, we compute the average of the three total correlation coefficients and use that as total correlation coefficient of a triplet. We report as final correlation coefficient the averages of the total correlation coefficient of each triplet over the 5 learned models for each method.

\begin{figure}[h!]
    \centering
    \includegraphics[scale = 0.5]{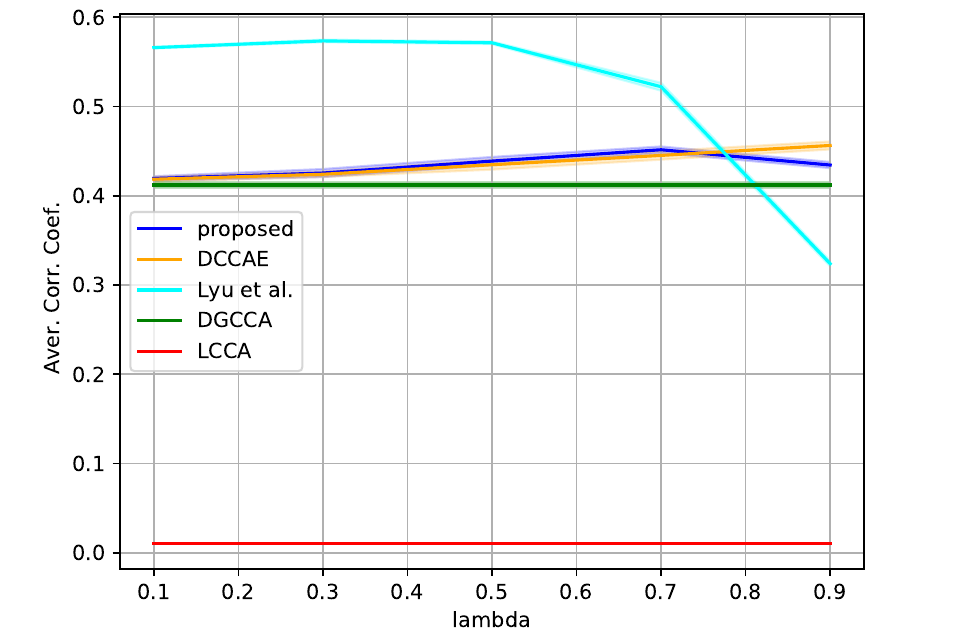}
    \caption{Average Correlation Coefficient for the XRMB dataset across different values of the $\lambda$ parameter.}
    \label{fig:Phon}
    \centering
    \includegraphics[scale = 0.45]{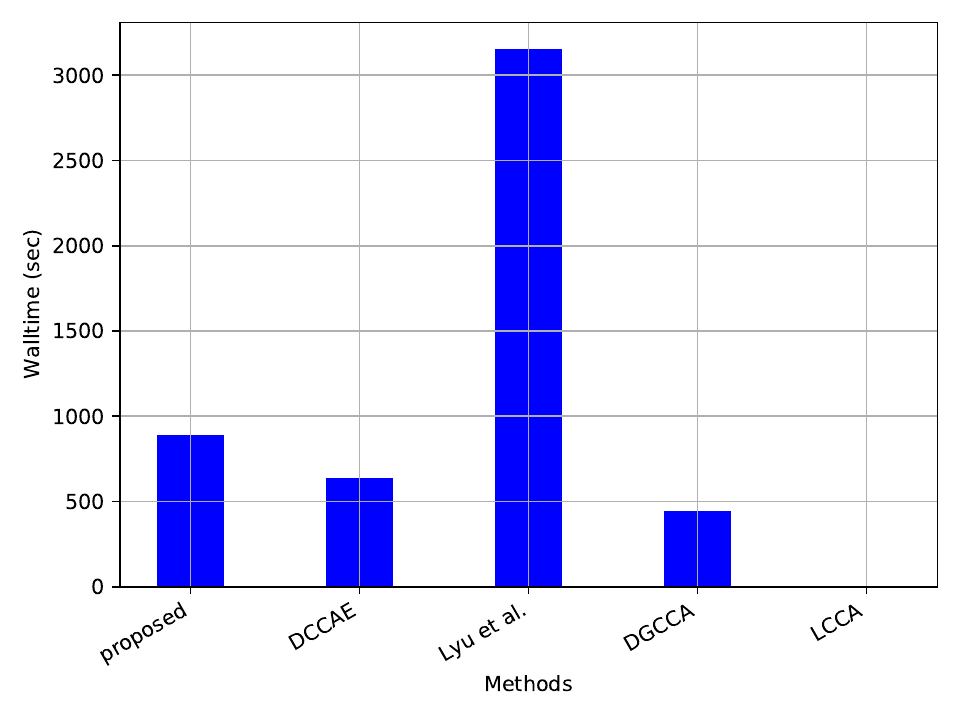}
    \caption{Average walltimes of all the considered methods for the XRMB dataset.}
    \label{fig:PhonTime}
    \vspace{-2mm}
\end{figure}

In Table \ref{tab:Phon}, we present the classification accuracy score that each method attained. Notice that MAX-VAR and DGCCA have no tuning parameter, hence we report their attained score in the fourth column ($\lambda=0.5$). In Fig. \ref{fig:Phon} we depict the total correlation coefficient as described above. The correlation coefficients for MAX-VAR and DGCCA are presented using straight lines. We observe that although our method does not attain the highest correlation coefficient, it achieves the highest classification accuracy. This observation comes to validate our statement that the quality of the embeddings of all the other methods can be deteriorated because of potential information leakage from the corresponding private components, but also because they open to capturing trivial solutions. As we see here, fictitious high correlation coefficients can be found by deep CCA-based methods, without however capturing useful information. At last, in Fig. \ref{fig:PhonTime}, we depict the average walltimes that emerged after training all the methods. We can see that the observations we noted in our experiments with the synthetic data carry on here, too, with the difference that the DCCAE method and the proposed method are more than three times faster than the method of Lyu et al. \cite{lyu2021understanding}.

\begin{table}
\centering
\caption{\label{tab:Phon}Average classification scores for the phoneme classification task across different values of the $\lambda$ parameter. \\}
\resizebox{.485\textwidth}{!}{%
\begin{tabularx}{1.268\linewidth}{|P{1.49cm}|P{1.42cm}|P{1.42cm}|P{1.42cm}|P{1.42cm}|P{1.42cm}|}
\hline
  & $\lambda=0.1$ & $\lambda=0.3$ & $\lambda=0.5$ & $\lambda=0.7$ & $\lambda=0.9$\\
 \hline
\text{MAX-VAR} & - & -  & 0.35$\pm$0 & - & - \\
\hline
\text{DGCCA}  & - & - & 0.46$\pm$0 & - & - \\
\hline
\text{DCCAE} &0.47$\pm$0.01 & 0.50$\pm$0.01 & 0.52$\pm$0.01 & 0.54$\pm$0.01 & 0.57$\pm$0 \\
\hline
Lyu et al. (2021) & 0.60$\pm$0.01 & 0.60$\pm$0.01 & 0.60$\pm$0 & 0.60$\pm$0 & 0.60$\pm$0\\
\hline
Proposed & 0.51$\pm$0.01 & 0.54$\pm$0.01 & 0.57$\pm$0.01 & 0.58$\pm$0.01 & \textbf{0.63}$\boldsymbol{\pm}$\textbf{0.01} \\
\hline
\end{tabularx}%
}
\vspace{-2mm}
\end{table}

\subsection{Multiview Noisy MNIST digits}

The Multiview Noisy MNIST dataset was introduced in \cite{wang2015deep} and it is a  popular dataset in multiview learning \cite{wang2015deep, lyu2021understanding}. It is based on the well-known MNIST dataset \cite{lecun1998gradient}, which consists of $28\times 28$ grayscale digit images, with 60,000 and 10,000 images for training and testing, respectively. Wang et al. \cite{wang2015deep} considered generating a two-view variation of the MNIST dataset, where the first view consists of randomly rotated digit images, while the second view consists of noisy digit images. More specifically, the dataset is generated as follows. The pixel values of all images of the initial dataset are first rescaled to the $\left[0,1\right]$ interval. Then, all the images are rotated by random angles uniformly sampled from $\left[-\pi/4, \pi/4\right]$ and the resulting images  are used to construct the first view. For each image of the first view, an image of the same digit (0-9) is randomly selected from the original dataset. Then, a noisy version of that image is obtained by adding noise to each pixel independently and uniformly sampled from $[0, 1]$. The pixel values of the resulting image are truncated to the $[0, 1]$ interval and the final image is used as the corresponding sample of the second view. At last, the original training set is further split into training and tuning sets of size 50,000 and 10,000, respectively. 

What is common between the realizations of the two views is the identity of the depicted digits. As a result, the encodings of the two views ideally should convey only information related to the identity of the pair. Hence, all the pairs sharing the same identity should have similar embeddings in the learned latent space. In this section, we compare the proposed method to all the state-of-the-art methods in the task of encoding the class label information. To achieve that, we follow \cite{wang2015deep, lyu2021understanding} and we consider a supervised setup and an unsupervised setup, as we did in our experiments with the synthetic dataset. For the supervised part, we test the performance of all the methods using a classifier, while for the unsupervised one, we measure the class separation in the learned latent space.

\begin{table}
\centering
\caption{\label{tab:Char}Network Architectures for the Multiview MNIST dataset.}
\vspace{2mm}
\resizebox{.485\textwidth}{!}{%
\begin{tabularx}{1.025\linewidth}{|P{3.8cm}|P{4.4cm}|}
\hline
Encoders ($\boldsymbol{f}_k$)  & Decoders ($\boldsymbol{w}_k$)\\
\hline
\hline
input: $28\times 28\times 1$ & input: latent\_dimension \\
$4\times 4$ Conv, 64, stride 2, ReLU  &  FC 256, ReLU  \\
$4\times 4$ Conv, 32, stride 2, ReLU &  FC $7\times 7 \times 32$, ReLU  \\
FC 256, ReLU &$4\times 4$ Conv\_Trans, 64 ReLU, stride 2   \\
FC latent\_dimension    & $4\times 4$ Conv\_Trans, 1, stride 2  \\
\hline
\end{tabularx}%
}
\vspace{-2mm}
\end{table}

Following \cite{wang2015deep, lyu2021understanding}, we consider a dimension equal to 10 for the common factor $\mathbf{g}$. In Table \ref{tab:Char} we present the architectures of the encoders and decoders that we used for our experiments with the Multiview MNIST dataset. The structures are common for the two views. In order to approximate the functions $\tau_k$ and $\phi_k$ of the method proposed in \cite{lyu2021understanding}, we use three hidden-layer neural networks with 64 neurons per layer, where the first two layers are followed by a ReLU activation function. For the training of all the models, we use the AdamW optimizer \cite{loshchilov2017decoupled} with initial learning rate of $0.001$ and a regularization parameter of $0.001$. We set the batch size equal to 500 samples and we run the algorithm for 40 epochs. For the method of \cite{lyu2021understanding}, we set the batch size for the maximization problem equal to $1000$ and the learning rate equal to $10^{-3}$. The dimensions of the individual components are set equal to 20 and 50, respectively, as it is mentioned in their experiments. Based on how this dataset is constructed, one can detect similarities with the synthetic dataset we considered above. In order to examine whether the proposed generative model or the generative model proposed by Lyu et al. \cite{lyu2021understanding} fits better on this dataset, we consider two different values for the regularization parameter for the independence promoting term, 0 (the independence promoting term is ignored) and 0.001 (the independence promoting term is considered), as we did in our experiments with the synthetic dataset. At last, we adopt the same early stopping policy as we did with the previous two datasets.

We train the models of all methods for $\lambda$ $\in$ $\left\{0.1, 0.3, 0.5, 0.7, 0.9\right\}$. For each value of $\lambda$ and method, we consider 5 different initializations resulting in 5 different learned models. For each learned model, we consider two tests corresponding to the supervised and the unsupervised evaluation setups, respectively, as we did with the synthetic dataset. To make the tests more challenging, we create three datasets in the latent space (training/validation/testing), by considering the embeddings of the samples of the noisy view for the three initial datasets (training/validation/testing), instead of the averages of the embeddings of the samples of both views. For the supervised test, we train a linear support vector machine (SVM) using the latent training data. The performance of the trained SVM is evaluated on the latent testing dataset and is measured by classification accuracy (CLA-ACC). Regarding the unsupervised test, we consider using K-means in order to split the latent training dataset into 10 clusters and evaluate how well the clusters agree with ground-truth labels. In the previous step, K-means is run 10 times with random initialization and the run with the best K-means objective is finally considered. We measure the clustering performance with three criteria, clustering accuracy (ACC), normalized mutual information (NMI), and adjusted Rand index (ARI) \cite{yeung2001details}. Finally, the final performance metrics of each method, for each value of $\lambda$, is determined by computing the average over the performance metrics obtained from all the 5 corresponding learned models.

\begin{figure}[t!]
    \centering
    \includegraphics[scale = 0.50]{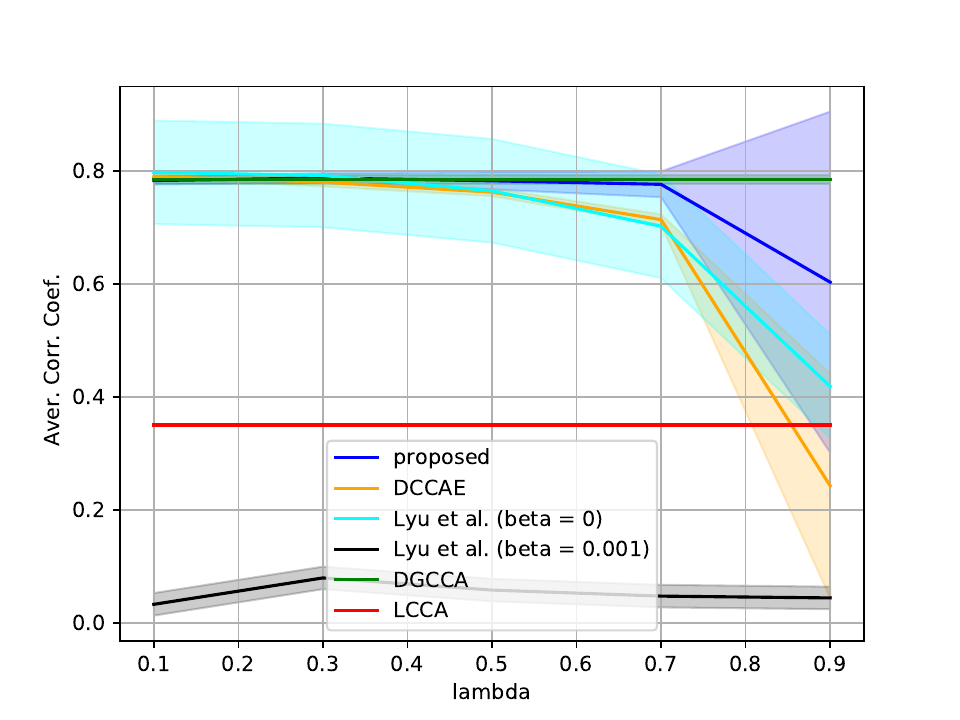}
    \caption{Average Correlation Coefficient for the Multiview dataset across different values of the $\lambda$ parameter.}
    \label{fig:Mnist}
    \vspace{2mm}
    \centering
    \includegraphics[scale = 0.44]{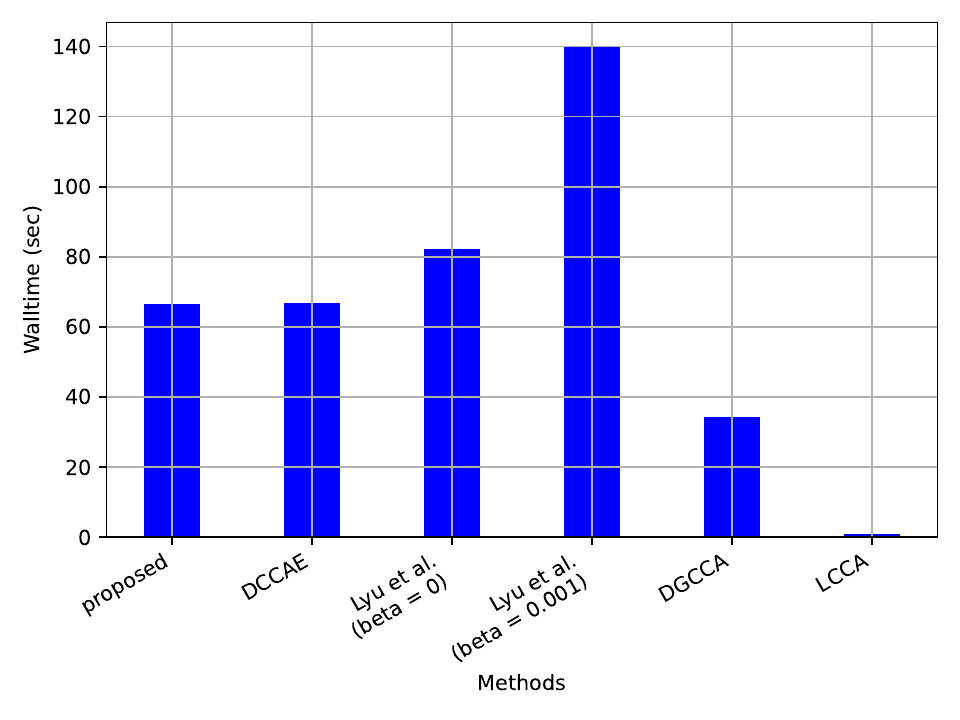}\vspace{-1mm}
    \caption{Average walltimes of all the considered methods for the Multiview MNIST dataset.}
    \label{fig:Mnist_time}
    \vspace{4mm}
    \end{figure}

In Tables \ref{tab:ACC}, \ref{tab:NMI}, \ref{tab:ARI}, and \ref{tab:CACC}, we present the final performance matrices of all the considered methods versus the considered values of the $\lambda$ parameter. We can observe that the proposed method achieves the best scores across all the considered metrics, consistently, and that also is the most robust to the variation of the $\lambda$ parameter. In Fig. \ref{fig:Mnist}, we present the average correlation coefficient that each method achieved over the 5 learned models for each value of the $\lambda$ parameter. We observe that the method proposed by \cite{lyu2021understanding} achieves a slightly higher correlation coefficient for small values of $\lambda$ and $\beta=0$, but it has significantly higher variance over the different initializations. On the other hand, the proposed method achieves comparably high correlation coefficients, with significantly less variance over the different initializations, and is less sensitive to the variation of the $\lambda$ parameter. Moreover, we can observe that all the performance scores and the correlation coefficient drop significantly when we consider the independence-promoting term for \cite{lyu2021understanding} method, which is an indication that the proposed generative model is more appropriate for this dataset compared to the one in \cite{lyu2021understanding}. In Fig. \ref{fig:Mnist_time}, we depict the walltime of all the methods for the Multiview Mnist dataset. We can see that the observations we noted for the previous two datasets carry on for the Multiview Mnist dataset, too. 

\begin{figure}[t!]
    \centering
     \centering
    \includegraphics[scale = 0.39]{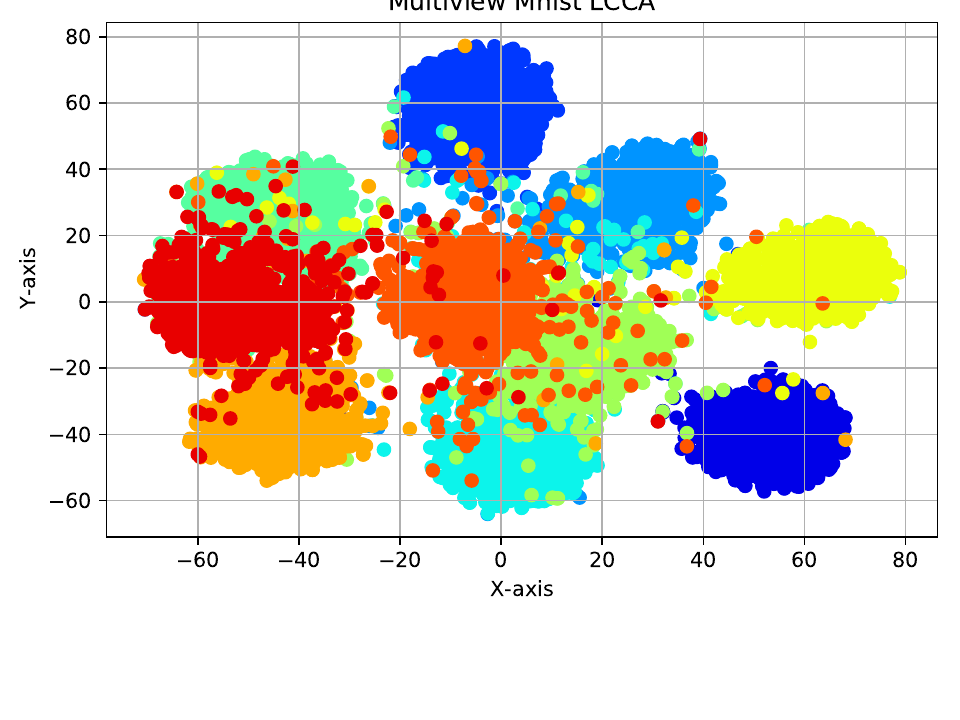}\vspace{-3mm}
    \includegraphics[scale = 0.39]{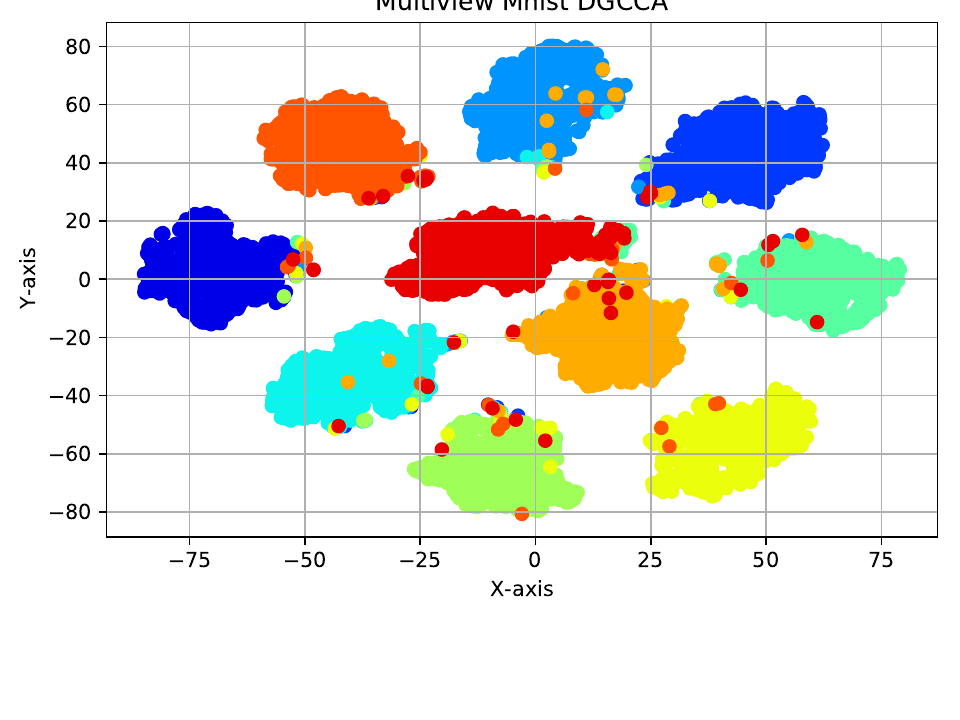}\vspace{-8mm}
    \caption{t-SNE visualizations of the embeddings of the Multiview MNIST dataset for the CCA (top) and the DGCCA (bottom) methods.}
    \label{fig:LCCAMnist}
    \vspace{-2mm}
\end{figure}

Next, we consider the t-SNE visualizations \cite{van2008visualizing} of the learned latent embeddings for the Multiview MNIST dataset. More specifically, we consider the samples appearing in the noisy view from the testing dataset and the learned models that achieved the lowest objective values, over the validation dataset, for all the methods. The resulting embeddings are fed to the t-SNE algorithm. In Fig. \ref{fig:LCCAMnist}, we present the obtained visualizations for the CCA and the DGCCA methods. In Fig. \ref{fig:DCCAEMnist}, we present the obtained visualizations for the DCCAE method over different values of $\lambda$. In Fig. \ref{fig:LyuMnist0} and Fig. \ref{fig:LyuMnist1}, we present the obtained visualizations for the method proposed in \cite{lyu2021understanding} over different values of $\lambda$, for $\beta=0$ and $\beta=0.001$, respectively. At last, in Fig. \ref{fig:XEMnist}, we present the obtained visualizations for the proposed method over different values of $\lambda$. We observe that the 10 clusters are well separated for all the deep methods for small values of $\lambda$, except for the method of Lyu et al. \cite{lyu2021understanding} and $\beta = 0.001$. In addition, we can observe that the proposed method is the only one that produces consistently clustered embeddings for all values of $\lambda$.

\begin{table}[t!]
\centering
\caption{\label{tab:ACC} Clustering accuracy (ACC) for the Multiview MNIST across different values of the $\lambda$ parameter. }
\resizebox{.485\textwidth}{!}{%
\begin{tabularx}{1.268\linewidth}{|P{1.49cm}|P{1.42cm}|P{1.42cm}|P{1.42cm}|P{1.42cm}|P{1.42cm}|}
\hline
  & $\lambda=0.1$ & $\lambda=0.3$ & $\lambda=0.5$ & $\lambda=0.7$ & $\lambda=0.9$\\
 \hline
\text{CCA} & - & -  & 0.89$\pm$0 & - & -\\
\hline
\text{DGCCA} & - & -  & 0.97$\pm$0 & - & - \\
\hline
\text{DCCAE}  &0.97$\pm$0 &0.97$\pm$0 &0.97$\pm$0 & 0.94$\pm$0.01 & 0.49$\pm$0.15\\
\hline
Lyu et al. (2021) $\beta=0$ &0.97$\pm$0& 0.97 $\pm$0& 0.94$\pm$0.05 &0.94$\pm$0.01& 0.62$\pm$0.11\\
\hline
Lyu et al. (2021) $\beta=0.001$ &0.21$\pm$0.02 & 0.22$\pm$0.02 &  0.22$\pm$0.02 & 0.21$\pm$0.01 & 0.21$\pm$0.02\\
\hline
Proposed & \textbf{0.98}$\pm$\textbf{0} &\textbf{0.98}$\pm$\textbf{0} &0.97$\pm$0 & 0.97$\pm$0 & 0.87$\pm$0.21\\
\hline
\end{tabularx}%
}\vspace{5mm}
\centering
\caption{\label{tab:NMI} Normalized mutual information (NMI) for the Multiview MNIST across different values of the $\lambda$ parameter. }
\resizebox{.485\textwidth}{!}{%
\begin{tabularx}{1.268\linewidth}{|P{1.49cm}|P{1.42cm}|P{1.42cm}|P{1.42cm}|P{1.42cm}|P{1.42cm}|}
\hline
  & $\lambda=0.1$ & $\lambda=0.3$ & $\lambda=0.5$ & $\lambda=0.7$ & $\lambda=0.9$\\
 \hline
\text{CCA}  & -  & - & 0.79$\pm$0 & - & -\\
\hline
\text{DGCCA} & - & - & 0.93$\pm$0 & - & - \\
\hline
\text{DCCAE} & \textbf{0.93}$\boldsymbol{\pm}$\textbf{0} & \textbf{0.93}$\boldsymbol{\pm}$\textbf{0} & 0.92$\pm$0 & 0.87$\pm$0.01 & 0.42$\pm$0.17\\
\hline
Lyu et al. (2021) $\beta=0$  &\textbf{0.93}$\pm$\textbf{0} & 0.92$\pm$0 & 0.90$\pm$0.03 & 0.87$\pm$0.01 & 0.56$\pm$0.11 \\
\hline
Lyu et al. (2021) $\beta=0.001$ &0.09$\pm$0.01 & 0.11$\pm$0.03 & 0.11$\pm$0.02 & 0.10$\pm$0.01 & 0.10$\pm$0.02\\
\hline
Proposed & \textbf{0.93}$\pm$\textbf{0} & \textbf{0.93}$\pm$\textbf{0} & \textbf{0.93}$\pm$\textbf{0.01} & \textbf{0.93}$\pm$\textbf{0} & 0.82$\pm$0.21\\
\hline
\end{tabularx}%
}\vspace{5mm}
\centering
\caption{\label{tab:ARI} Adjusted Rand index (ARI) for the Multiview MNIST dataset across different values of the $\lambda$ parameter. }
\resizebox{.485\textwidth}{!}{%
\begin{tabularx}{1.268\linewidth}{|P{1.49cm}|P{1.42cm}|P{1.42cm}|P{1.42cm}|P{1.42cm}|P{1.42cm}|}
\hline
  & $\lambda=0.1$ & $\lambda=0.3$ & $\lambda=0.5$ & $\lambda=0.7$ & $\lambda=0.9$\\
 \hline
\text{CCA}  & - & - & 0.78$\pm$0 & - & - \\
\hline
\text{DGCCA} & -  & -  & 0.94$\pm$0  & -  & - \\
\hline
\text{DCCAE} & \textbf{0.95}$\boldsymbol{\pm}$\textbf{0} & 0.94$\pm$0 & 0.93$\pm$0 & 0.88$\pm$0.01 & 0.33$\pm$0.16 \\
\hline
Lyu et al. (2021) $\beta=0$ &0.94$\pm$0 & 0.93$\pm$0 & 0.90$\pm$0.05 & 0.87$\pm$0.02 & 0.48$\pm$0.12 \\
\hline
Lyu et al. (2021) $\beta=0.001$ &0.04$\pm$0.01 & 0.06$\pm$0.02 & 0.06$\pm$0.01 & 0.05$\pm$0.01 & 0.05$\pm$0.02\\
\hline
Proposed & \textbf{0.95}$\boldsymbol{\pm}$\textbf{0} & \textbf{0.95}$\boldsymbol{\pm}$\textbf{0} & \textbf{0.95}$\boldsymbol{\pm}$\textbf{0.01} & 0.94$\pm$0 & 0.81$\pm$0.26\\
\hline
\end{tabularx}%
}\vspace{5mm}
\centering
\caption{\label{tab:CACC} Classification accuracy (CLA-ACC) for the Multiview MNIST across different values of the $\lambda$ parameter. }
\resizebox{0.485\textwidth}{!}{%
\begin{tabularx}{1.268\linewidth}{|P{1.49cm}|P{1.42cm}|P{1.42cm}|P{1.42cm}|P{1.42cm}|P{1.42cm}|}
\hline
  & $\lambda=0.1$ & $\lambda=0.3$ & $\lambda=0.5$ & $\lambda=0.7$ & $\lambda=0.9$\\
 \hline
\text{CCA}  & - & - & 0.91$\pm$0 & - & -  \\
\hline
\text{DGCCA}  & - & - & 0.97$\pm$0 & - & - \\
\hline
\text{DCCAE} & \textbf{0.98}$\boldsymbol{\pm}$\textbf{0} &0.97$\pm$0 &0.97$\pm$0 & 0.95$\pm$0.01 & 0.66$\pm$0.12 \\
\hline
Lyu et al. (2021) $\beta=0$ &0.97$\pm$0 & 0.97$\pm$0 & 0.96$\pm$0.01 & 0.95$\pm$0.01 & 0.82$\pm$0.05 \\
\hline
Lyu et al. (2021) $\beta=0.001$ &0.38$\pm$0.05 & 0.39$\pm$0.05 & 0.44$\pm$0.03 & 0.41$\pm$0.03 & 0.39$\pm$0.02\\
\hline
Proposed & \textbf{0.98}$\boldsymbol{\pm}$\textbf{0} &\textbf{0.98}$\boldsymbol{\pm}$\textbf{0} &0.97$\pm$0 & 0.97$\pm$0 & 0.94$\pm$0.07 \\
\hline
\end{tabularx}%
}
\end{table}


\section{CONCLUSIONS}
In this work, we revisited the problem of DGCCA and we highlighted the advantages and limitations that the previous works have. We proposed a novel formulation of the problem that overcomes most of those limitations and an algorithm for estimating the common latent random vector given jointly observed views. We tested the proposed concepts in artificial and real-life datasets. The obtained results corroborate our claims that the proposed framework is more appropriate and performs better in a series of tasks. 

\appendix

\section*{Estimating random vector $\mathbf{g}$ - Further discussion}

We estimate random vector $\mathbf{g}$ from realizations of random vectors $\mathbf{x}^{(k)}$ via the conditional expectation 
\begin{equation}
    \mathbb{E}\left[\mathbf{g}\big|\mathbf{x}^{(1)},\ldots,\mathbf{x}^{(K)}\right].
    \label{CondExp1}
\end{equation}
Although this estimator is optimal under the Mean Squared Error criterion, it comes with two important practical caveats. First, it is a very high-dimensional and generally unstructured function of all random vectors $\mathbf{x}^{(1)},\ldots,\mathbf{x}^{(K)}$, the approximation of which can be challenging in terms of complexity and scalability. The second caveat comes into play when we would also like to be able to estimate $\mathbf{g}$ from partial realizations of  $\left\{\mathbf{x}^{(k)}\right\}_{k\in\mathbb{S}}$, with $\mathbb{S}\subset\left[K\right]$ -- which is often important in practice. Since the optimal estimators, in this case, are functions of the one in (\ref{CondExp1}) via
\begin{equation}
\small
    \mathbb{E}\left[\mathbf{g}\big|\left\{\mathbf{x}^{(k)}\right\}_{k\in\mathbb{S}}\right]= \mathbb{E}_{\left\{\mathbf{x}^{(k)}\right\}_{k\not\in\mathbb{S}}}\left[\mathbb{E}\left[\mathbf{g}\big|\mathbf{x}^{(1)},\ldots,\mathbf{x}^{(K)}\right]\right],
    \label{ParCondExp1}
\end{equation}
we can see that given the estimator in (\ref{CondExp1}), the problem of computing the optimal estimators with respect to the subset $\mathbb{S}$ boils down to finding ways to efficiently compute different expectations of (\ref{CondExp1}) over the views that belong to the complement of $\mathbb{S}$. 

Considering the extreme case where each $\mathbb{S}$ is a singleton, we may consider combining multiple single-view estimators of ${\bf g}$, each of them still given by (\ref{ParCondExp1}). Although each of these estimators may have significantly poorer prediction capabilities compared to the one in (\ref{CondExp1}), they have the advantage that they may allow designing efficient and scalable frameworks for estimating $\mathbf{g}$ by combining the single view based estimates from each random vector $\mathbf{x}^{(k)}$. The problem that arises here is how may one combine such estimators in a way that is efficient, consistent, and also enables the estimation of $\mathbf{g}$ even from a single view. The first formulation of DGCCA falls along those lines, since it aims at finding single view-based estimators of $\mathbf{g}$ that are in agreement, and hence potentially suboptimal but consistent with one another. 

\section*{Proof of Theorem 1}

\begin{customthm}{1}
    Under the proposed generative model in (\ref{GenModel}), consider a solution of the optimization problem in (\ref{Prop_1}). If the following additional assumptions,
    \begin{itemize}
        \item[(i)] functions $\boldsymbol{v}_k$ are also partially invertible w.r.t. $\mathbf{g}$, i.e., there exist functions $\boldsymbol{u}_k:\mathbb{R}^{D_k}\rightarrow\mathbb{R}^F$, such that $\boldsymbol{u}_k\left(\mathbf{x}^{(k)}\right)=\mathbf{g}$, for all $\mathbf{x}^{(k)}$ and $k\in\left[K\right]$,
         \item[(ii)] there exists mean dependence between at least one pair of random vectors $\mathbf{g}$ and $\mathbf{x}^{(k')}$, i.e. $$\mathbb{E}\left[\mathbf{x}^{(k')}|\mathbf{g}\right]\neq \mathbb{E}\left[\mathbf{x}^{(k')}\right] \text{ for a } k'\in\left[K\right] \text{and all } \mathbf{g}\in\mathbb{R}^F,$$
    \end{itemize}
    hold, then the learned encodings $\boldsymbol{f}_k\left(\mathbf{x}^{(k)}\right)$ have to be non trivial functions only of $\mathbf{g}$, i.e. $\boldsymbol{f}_k\left(\boldsymbol{v}_{k}\left(
    \begin{bmatrix}
    \mathbf{g}\\
    \mathbf{c}^{(k)}
    \end{bmatrix}
    \right)\right)=\boldsymbol{\gamma}\left(\mathbf{g}\right)$, where $\boldsymbol{\gamma}:\mathbb{R}^F\rightarrow\mathbb{R}^F$. Moreover, if
    \begin{itemize}
        \item[(iii)]the conditional expectation,
$\mathbb{E}\left[ \begin{bmatrix}\mathbf{x}^{(1)^T}, \ldots, \mathbf{x}^{(K)^T}\end{bmatrix}^T \big|\mathbf{g}\right]$, is an invertible function of $\mathbf{g}$,
    \end{itemize}then function $\boldsymbol{\gamma}$ is also invertible and the latent common random vector $\mathbf{g}$ is identifiable up to invertible nonlinearities.
\end{customthm}

\begin{proof}

We begin by considering the function 
$$\mathcal{R}^{(K)}\left(\boldsymbol{f}_1,...,\boldsymbol{f}_K,\hat{\mathbf{g}}\right):=\sum_{k=1}^K\mathbb{E}\left[\left\|\boldsymbol{f}_k\left(\mathbf{x}^{(k)}\right)-\hat{\mathbf{g}}\right\|^2\right].$$
Because of the generative model in (\ref{GenModel}) and additional assumption (i), we have that the level set 
\begin{equation}
     \mathcal{S}_0:=\left\{\left(\left\{\boldsymbol{f}_k\right\}_{k=1}^K,\hat{\mathbf{g}}\right):\mathcal{R}^{(K)}\left(\boldsymbol{f}_1,...,\boldsymbol{f}_K,\hat{\mathbf{g}}\right)=0\right\}
\end{equation} 
is non empty, since $\left(\left\{\boldsymbol{u}_k\right\}_{k=1}^K, \mathbf{g}\right)\in\mathcal{S}_0$. Moreover, notice that for any function $\boldsymbol{\gamma}:\mathbb{R}^F\rightarrow\mathbb{R}^F$, also 
$\left(\left\{\boldsymbol{\gamma}\circ\boldsymbol{u}_k\right\}_{k=1}^K,\boldsymbol{\gamma}\left(\mathbf{g}\right)\right)\in\mathcal{S}_0$. Now, let $\left(\left\{\bar{\boldsymbol{f}}_k\right\}_{k=1}^K, \bar{\mathbf{g}}\right)$ be an element of $\mathcal{S}_0$ and functions $\bar{\boldsymbol{h}}_k:\mathbb{R}^{F+L_k}\rightarrow\mathbb{R}^F$ be defined according to
$$\bar{\boldsymbol{h}}_k\left(\begin{bmatrix}
   \mathbf{g}\\ \mathbf{c}^{(k)} 
\end{bmatrix}\right):=\bar{\boldsymbol{f}}_k\left(\boldsymbol{v}_k\left(\begin{bmatrix}
   \mathbf{g}\\ \mathbf{c}^{(k)} 
\end{bmatrix}\right)\right).$$
For any $k_1, k_2\in\left[K\right]$, with $k_1\neq k_2$, we have
that for any realization of $\mathbf{g}$, $\mathbf{g}'\in\mathbb{R}^F$, it holds that
\begin{equation*}
    \bar{\boldsymbol{h}}_{k_1}\left(\begin{bmatrix}
   \mathbf{g}'\\ \mathbf{c}^{(k_1)} 
\end{bmatrix}\right)=\bar{\boldsymbol{h}}_{k_2}\left(\begin{bmatrix}
   \mathbf{g}'\\ \mathbf{c}^{(k_2)}
\end{bmatrix}\right),
\end{equation*}
for all $\mathbf{c}^{(k_1)}\in\mathbb{R}^{L_{k_1}}$ and $\mathbf{c}^{(k_2)}\in\mathbb{R}^{L_{k_2}}$. Hence functions $\left\{\bar{\boldsymbol{h}}_k\right\}_{k=1}^K$ have to be functions only of $\mathbf{g}$, since, for any fixed $\mathbf{g}'$, the equalities hold no matter what the values of $\mathbf{c}^{(k_1)}$ and $\mathbf{c}^{(k_2)}$ are. In other words, there exists a function $\bar{\boldsymbol{\gamma}}: \mathbf{R}^F\rightarrow\mathbf{R}^F$, such that
\begin{equation*}
  \bar{\boldsymbol{h}}_{k}\left(\begin{bmatrix}
   \mathbf{g}'\\ \mathbf{c}^{(k)} 
\end{bmatrix}\right) = \bar{\boldsymbol{\gamma}}\left(\mathbf{g}'\right), \text{ for all } \mathbf{g}'\in\mathbb{R}^F \text{ and } k\in\left[K\right].
\end{equation*}
Therefore, $\bar{\mathbf{g}}$ must be equal to $\bar{\boldsymbol{\gamma}}\left(\mathbf{g}\right)$. Since the above arguments hold for any feasible $\bar{\mathbf{g}}$, it follows that the elements of $\mathcal{S}_0$ are given by $\left(\left\{\boldsymbol{\gamma}\circ\boldsymbol{u}_k\right\}_{k=1}^K, \boldsymbol{\gamma}\left(\mathbf{g}\right)\right)$, where  $\boldsymbol{\gamma}$ can be any function $\mathbb{R}^F\rightarrow\mathbb{R}^F$.

Now, let us focus on the objective of (\ref{Prop_1}),
$$\mathcal{B}^{(K)}\left(\boldsymbol{w}_1,\ldots,\boldsymbol{w}_K,\hat{\mathbf{g}}\right) = \sum_{k=1}^K\mathbb{E}\left[\left\|\boldsymbol{w}_k\left(\hat{\mathbf{g}}\right)-\mathbf{x}^{(k)}\right\|^2\right].$$ The only dependency between the objective and the constraints of (\ref{Prop_1}) is through random variable $\hat{\mathbf{g}}$. On the other hand, as we showed above, a feasible random vector $\hat{\mathbf{g}}$ has to be a function of the common latent random vector $\mathbf{g}$. Given that, the optimization problem of (\ref{Prop_1}) can be written as
\begin{equation}
\small
\begin{split}
    \min_{ \boldsymbol{\gamma}, \left\{\boldsymbol{w}_k\in\mathcal{C}_w\right\}_{k=1}^K}&\sum_{k=1}^K\mathbb{E}\left[\left\|\boldsymbol{w}_k\left(\boldsymbol{\gamma}\left(\mathbf{g}\right)\right)-\mathbf{x}^{(k)}\right\|^2\right]\\
\text{s.t.}\hspace{7.5mm}&\hspace{-4mm}\mathbb{E}\left[\boldsymbol{\gamma}\left(\mathbf{g}\right)\boldsymbol{\gamma}\left(\mathbf{g}\right)^T\right]=\mathbf{I}_F,\hspace{2mm}\mathbb{E}\left[\boldsymbol{\gamma}\left(\mathbf{g}\right)\right]=\mathbf{0}_F
    \end{split}
    \label{Prop_13}
    \raisetag{28pt}
\end{equation}
As long as the distribution of $\mathbf{g}$, $\mathcal{D}_{\mathbf{g}}$, is non-singular, the existence of an invertible function $\boldsymbol{\gamma}$, under which $\mathbb{E}\left[\boldsymbol{\gamma}\left(\mathbf{g}\right)\boldsymbol{\gamma}\left(\mathbf{g}\right)^T\right]=\mathbf{I}_F$ and $\mathbb{E}\left[\boldsymbol{\gamma}\left(\mathbf{g}\right)\right]=\mathbf{0}_F,$ emerges naturally after considering whitening transformations. Therefore, the feasible sets of problems (\ref{Prop_1}) and (\ref{Prop_13}) are nonempty.

Given a function $\boldsymbol{\gamma}$, the conditionally optimal functions $\left\{\tilde{\boldsymbol{w}}_k\right\}_{k=1}^K$
can be obtained by $\tilde{\boldsymbol{w}}_k\left(\boldsymbol{\gamma}\left(\mathbf{g}\right)\right)=\mathbb{E}\left[\mathbf{x}^{(k)}|\boldsymbol{\gamma}\left(\mathbf{g}\right)\right].$
Because of our second additional assumption, optimal functions $\left\{\boldsymbol{w}_k\circ\boldsymbol{\gamma}\right\}_{k=1}^K$ have to be non trivial functions in the sense that they have to be nonconstant. To see that, notice that letting $\boldsymbol{w}_{k'}\left(\boldsymbol{\gamma}\left(\mathbf{g}\right)\right)=\mathbb{E}\left[\mathbf{x}^{(k')}\right]$ would induce a strictly higher MSE than $\boldsymbol{w}_{k'}\left(\boldsymbol{\gamma}^*\left(\mathbf{g}\right)\right)=\mathbb{E}\left[\mathbf{x}^{(k')}|\boldsymbol{\gamma}^*\left(\mathbf{g}\right)\right]=\mathbb{E}\left[\mathbf{x}^{(k')}|\mathbf{g}\right]$, which is the best MSE predictor of $\mathbf{x}^{(k')}$ over all possible functions $\boldsymbol{\gamma}$. Non triviality of all optimal functions $\boldsymbol{w}_{k'}\circ\boldsymbol{\gamma}$ implies non triviality of all optimal functions $\boldsymbol{\gamma}:\mathbb{R}^F\rightarrow\mathbb{R}^F$, which have to satisfy the following three properties
\begin{itemize}
\item[(a)]$\mathbb{E}\left[\mathbf{x}^{(k)}|\boldsymbol{\gamma}\left(\mathbf{g}\right)\right]=\mathbb{E}\left[\mathbf{x}^{(k)}|\mathbf{g}\right], ~\forall k\in\left[K\right]$,
    \item[(b)] $\mathbb{E}\left[\boldsymbol{\gamma}\left(\mathbf{g}\right)\boldsymbol{\gamma}\left(\mathbf{g}\right)^T\right]=\mathbf{I}_F$,
    \item[(c)] $ \mathbb{E}\left[\boldsymbol{\gamma}\left(\mathbf{g}\right)\right]=\mathbf{0}_F$.
\end{itemize}

Let $\boldsymbol{\phi}:\mathbb{R}^F\rightarrow\mathbb{R}^{\sum_{k=1}^KD_k}$ be given by the conditional expectation
$\boldsymbol{\phi}\left(\mathbf{g}\right) := \mathbb{E}\left[  \begin{bmatrix}\mathbf{x}^{(1)^T}, \ldots, \mathbf{x}^{(K)^T}\end{bmatrix}^T \big| \mathbf{g}\right].$ Under the third additional assumption, there exists a function $\boldsymbol{\psi}:\mathbb{R}^{\sum_{k=1}^KD_k}\rightarrow\mathbb{R}^F$, such that 
$\boldsymbol{\psi}\left(\boldsymbol{\phi}\left(\mathbf{g}\right)\right) = \mathbf{g}, \text{ for all }\mathbf{g}\in\mathbb{R}^F.$
For an optimal collection of functions $\left\{\boldsymbol{w}^*_k\right\}_{k=1}^K$ and $\boldsymbol{\gamma}^*$ c.f. (\ref{Prop_1}), let $\boldsymbol{\phi}^*\left(\boldsymbol{\gamma}^*\left(\mathbf{g}\right)\right):=\begin{bmatrix}
        \boldsymbol{w}^*_1\left(\boldsymbol{\gamma}^*\left(\mathbf{g}\right)\right)^T,
        \ldots,
        \boldsymbol{w}^*_K\left(\boldsymbol{\gamma}^*\left(\mathbf{g}\right)\right)^T
    \end{bmatrix}^T.$
Because of property (a), we have that 
\begin{equation}
\boldsymbol{\phi}\left(\mathbf{g}\right)=\boldsymbol{\phi}^*\left(\boldsymbol{\gamma}^*\left(\mathbf{g}\right)\right), \text{ for all }\mathbf{g}\in\mathbb{R}^F,
\end{equation}
which implies that an optimal function $\boldsymbol{\gamma}^*$ has an inverse function, given by $\boldsymbol{\gamma}^{*^{-1}} = \boldsymbol{\psi}\circ\boldsymbol{\phi}^*$, since
\begin{equation}
    \mathbf{g} = \boldsymbol{\psi}\left(\boldsymbol{\phi}^*\left(\boldsymbol{\gamma}^*\left(\mathbf{g}\right)\right)\right), \text{ for all }\mathbf{g}\in\mathbb{R}^F.
\end{equation}
As a result, under the additional assumption (iii), the latent common random vector is identifiable up to invertible nonlinearities expressed through $\boldsymbol{\gamma}^*$.

\end{proof}

\bibliographystyle{IEEEtran}
\bibliography{References}

\begin{IEEEbiography}[{\includegraphics[scale=0.2, angle=-90, clip, trim = {14cm 9.5cm 12cm 10cm}]{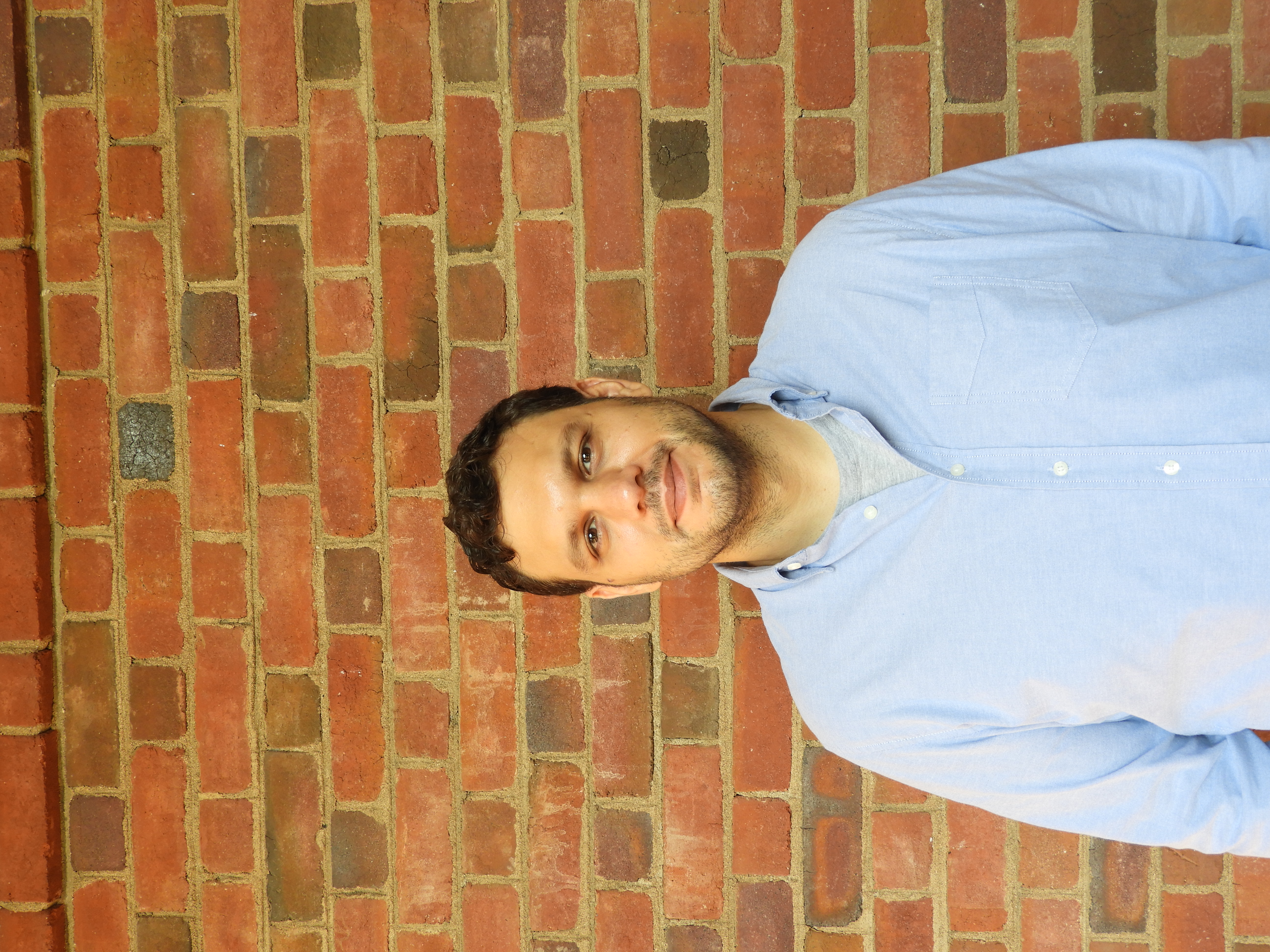}}]{Paris A. Karakasis}received the Diploma and M.Sc. degree in Electrical and Computer Engineering from the Technical University of Crete, Chania, Greece, in 2017 and 2019, respectively. Currently, he is a Ph.D. student at the Electrical and Computer Engineering Department, University of Virginia, Charlottesville, VA, USA. His research interests include signal processing, optimization, machine learning, tensor decomposition, and graph mining.
\end{IEEEbiography}

\begin{IEEEbiography}[{\includegraphics[scale=0.15]{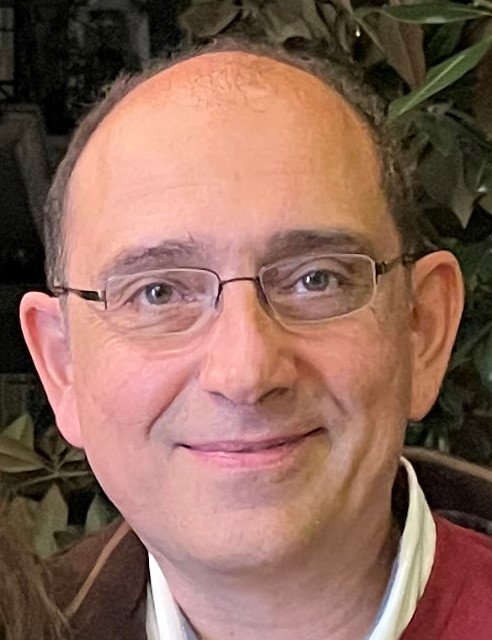}}]
{Nicholas D. Sidiropoulos} (Fellow, IEEE) received the Diploma in electrical engineering from the Aristotle University of Thessaloniki, Thessaloniki, Greece, and the M.S. and Ph.D. degrees in electrical engineering from the University of Maryland at College Park, College Park, MD, USA, in 1988, 1990, and 1992, respectively. He is the Louis T. Rader Professor in the Department of ECE at the University of Virginia. He has previously served on the faculty at the University of Minnesota and the Technical University of Crete, Greece. His research interests are in signal processing, communications, optimization, tensor decomposition, and machine learning. He received the NSF/CAREER award in 1998, the IEEE Signal Processing Society (SPS) Best Paper Award in 2001, 2007, 2011, and 2023, and the IEEE SPS Donald G. Fink Overview Paper Award in 2023. He served as IEEE SPS Distinguished Lecturer (2008–2009), Vice President - Membership (2017–2019) and Chair of the SPS Fellow evaluation committee (2020-2021). He received the 2010 IEEE SPS Meritorious Service Award, the 2013 Distinguished Alumni Award of the ECE Department of University of Maryland, the 2022 EURASIP Technical Achievement Award, and the 2022 IEEE SPS Claude Shannon - Harry Nyquist Technical Achievement Award. He is a fellow of EURASIP (2014).
\end{IEEEbiography}

\clearpage
\begin{figure}[t]
    \centering
    \begin{minipage}{.5\textwidth}
    \includegraphics[scale = 0.39]{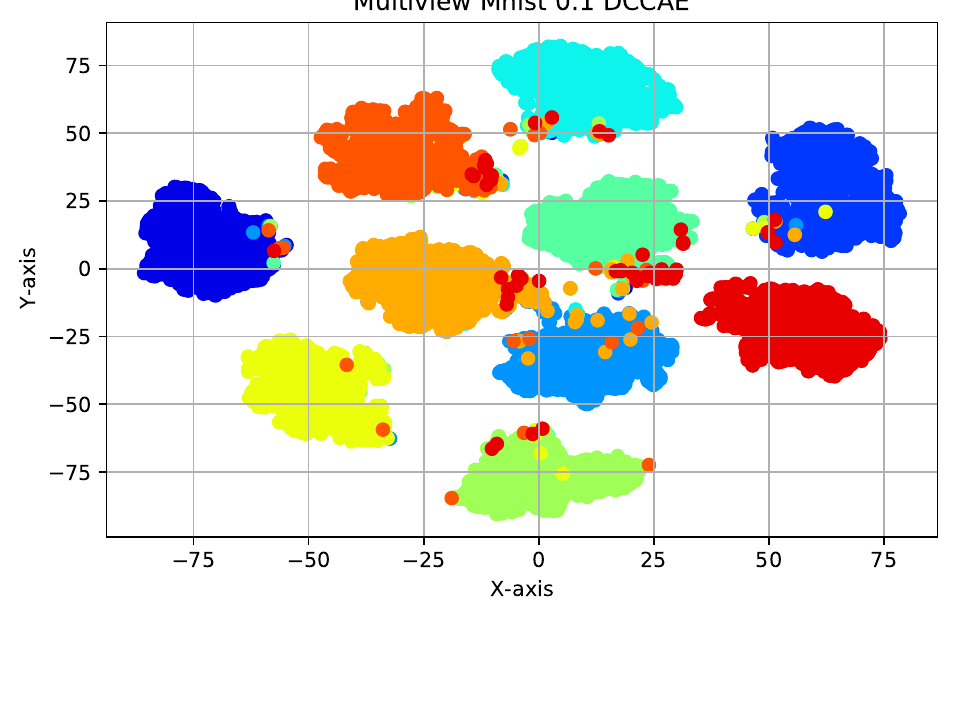}
    \includegraphics[scale = 0.39]{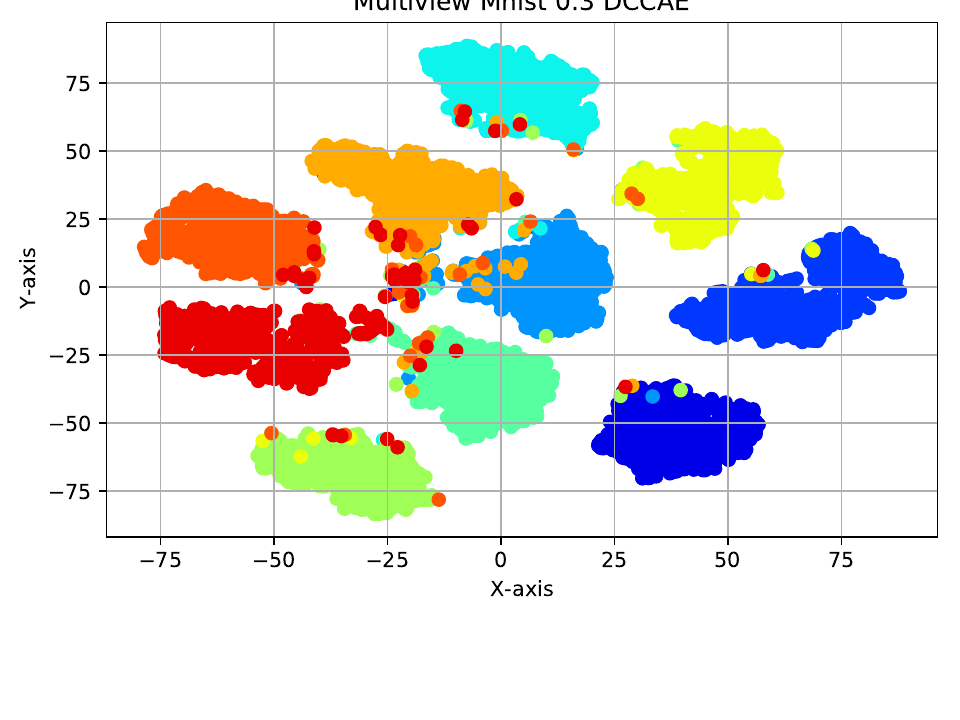}
    \includegraphics[scale = 0.39]{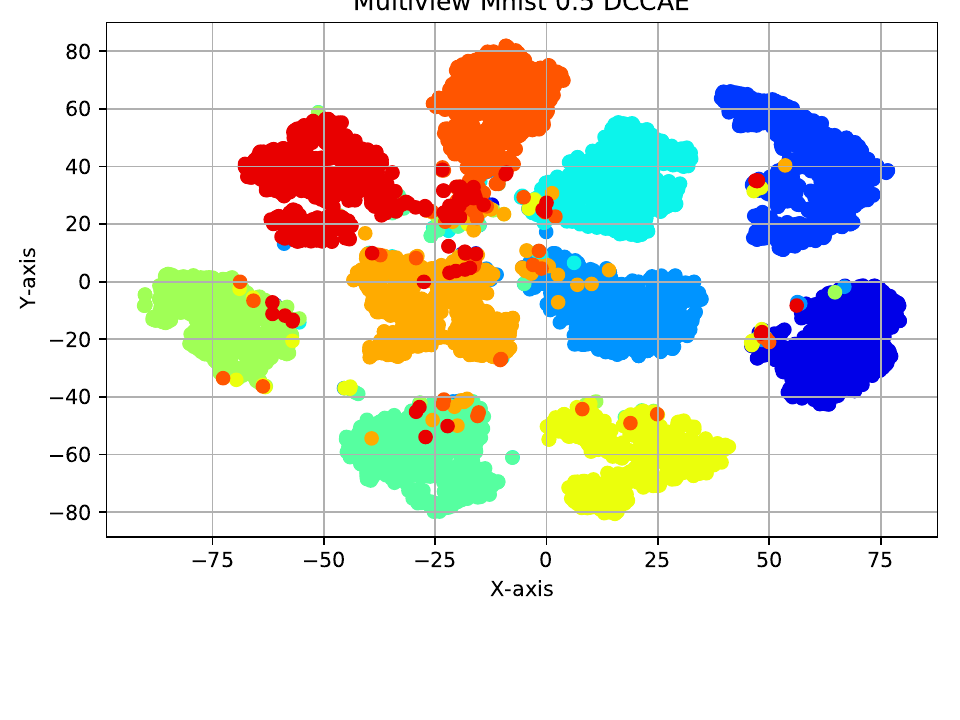}
    \includegraphics[scale = 0.39]{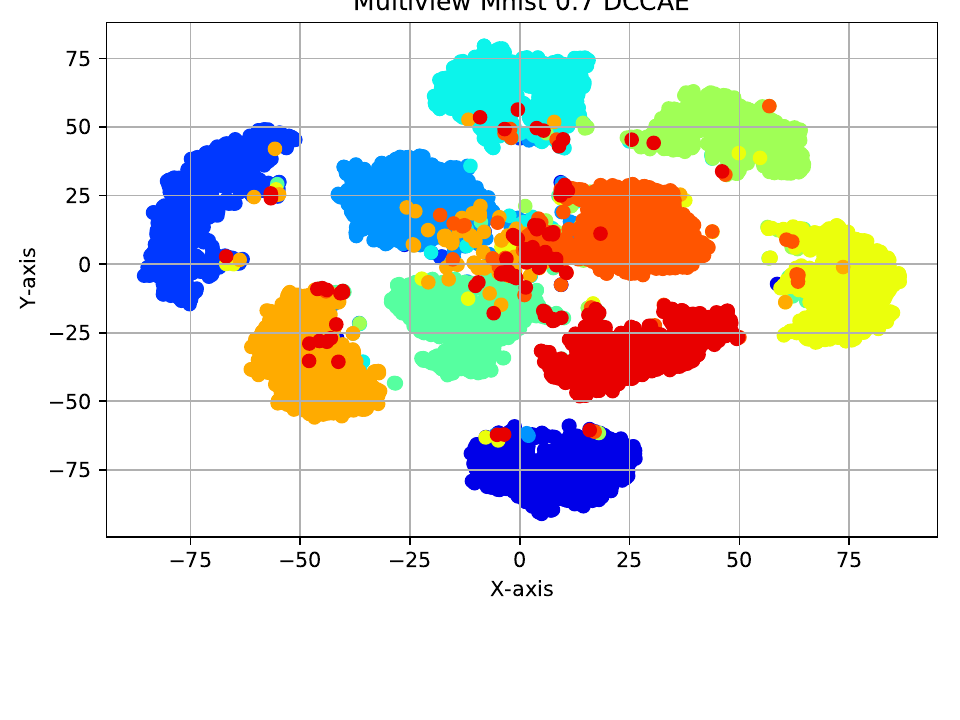}
    \includegraphics[scale = 0.39]{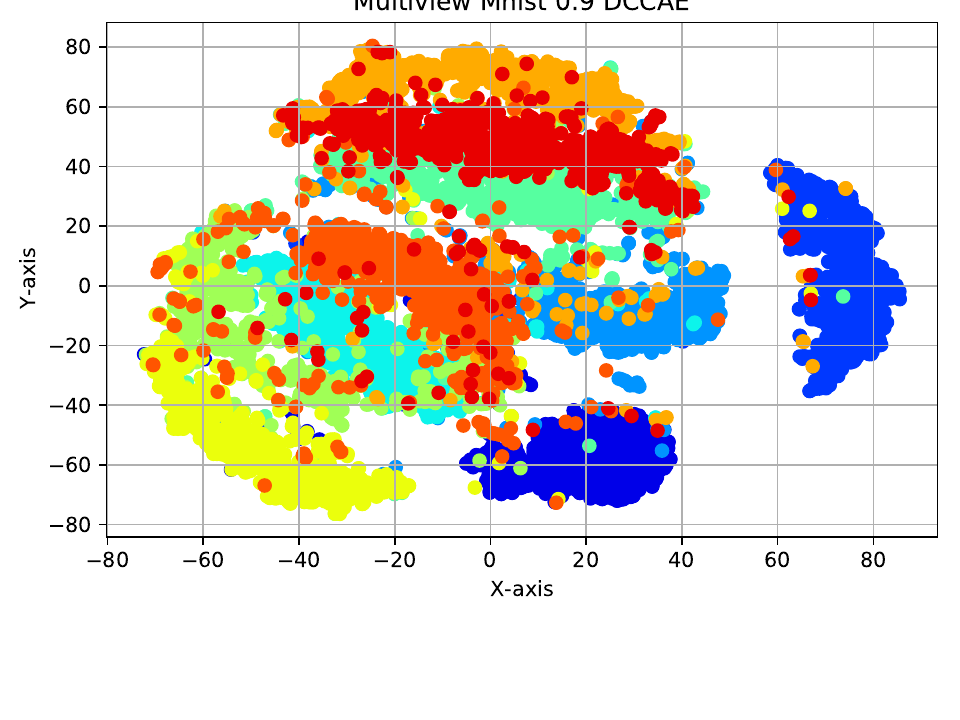}
    \vspace*{-6mm}
    \caption{t-SNE visualizations of the embeddings of the\\ Multiview MNIST dataset for the DCCAE method for\\ different values of $\lambda$ ($\lambda=0.1$-top to $\lambda=0.9$-bottom).\phantom{xxx}}
    \label{fig:DCCAEMnist}
    \end{minipage}%
    \begin{minipage}{.5\textwidth}
    \centering
    \includegraphics[scale = 0.39]{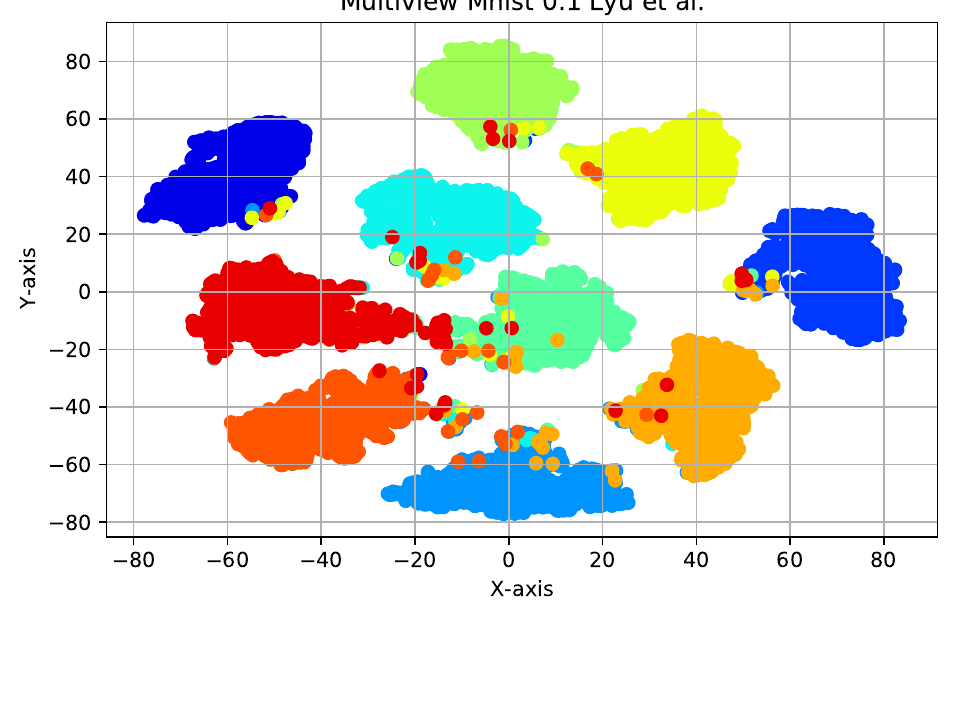}
    \includegraphics[scale = 0.39]{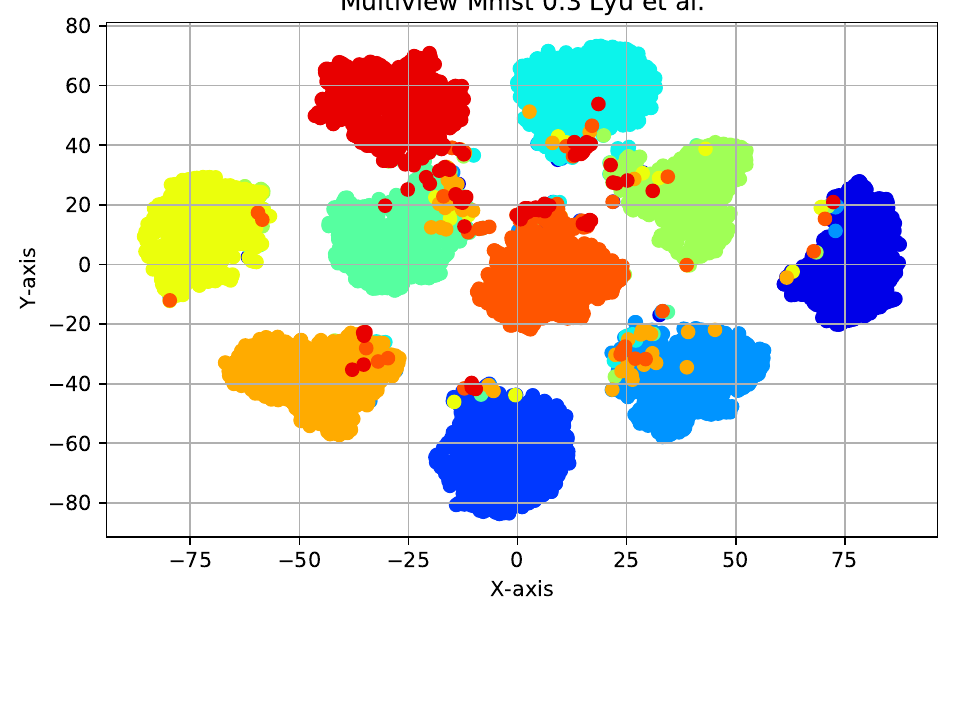}
    \includegraphics[scale = 0.39]{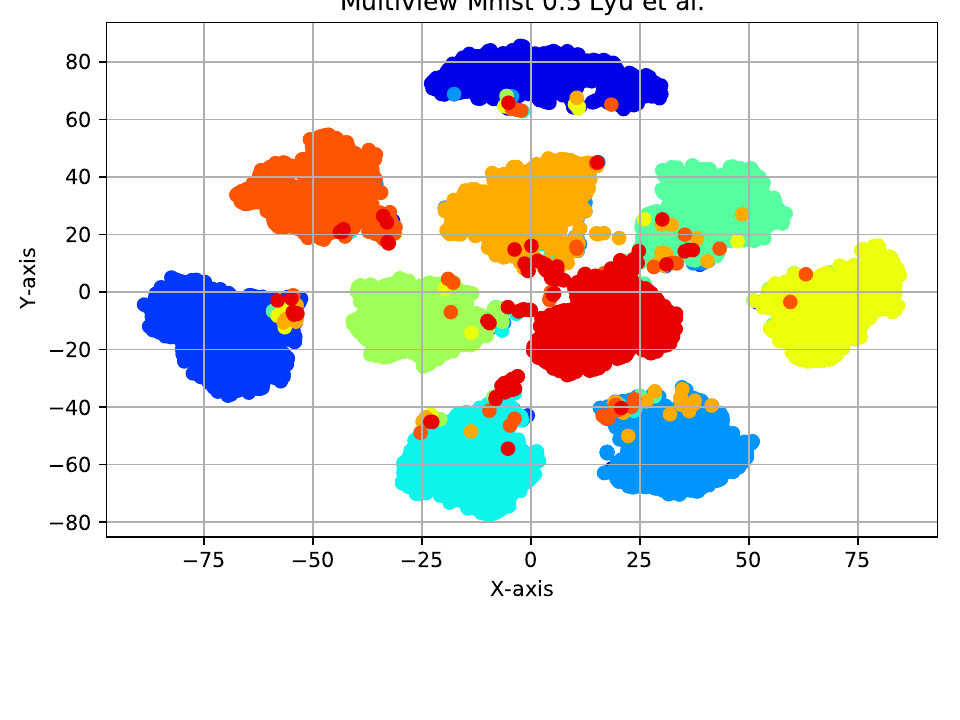}
    \includegraphics[scale = 0.39]{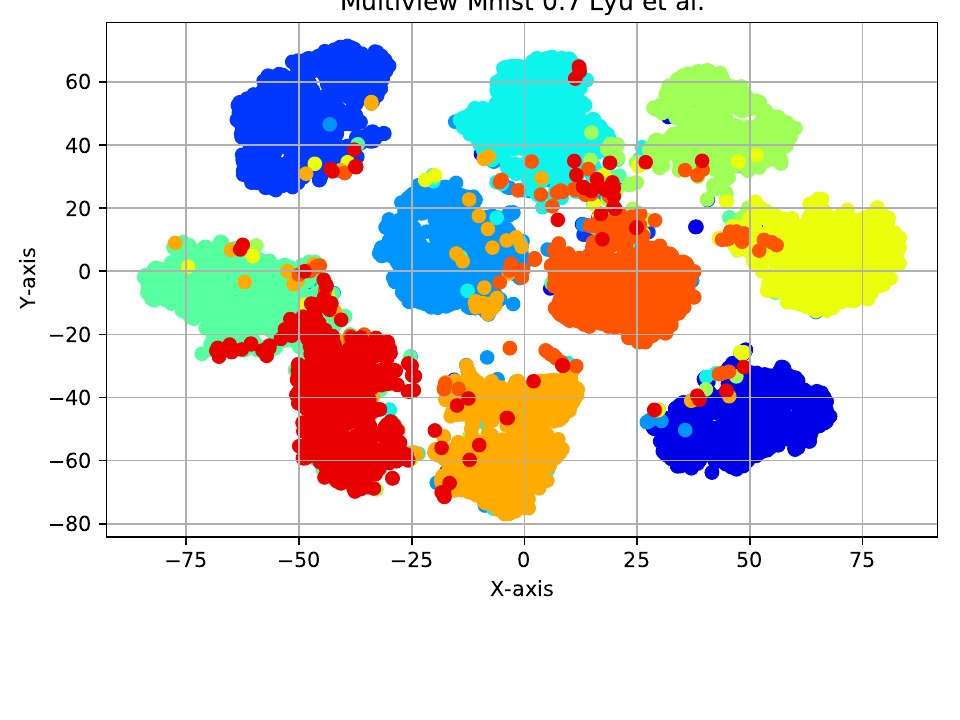}
    \includegraphics[scale = 0.39]{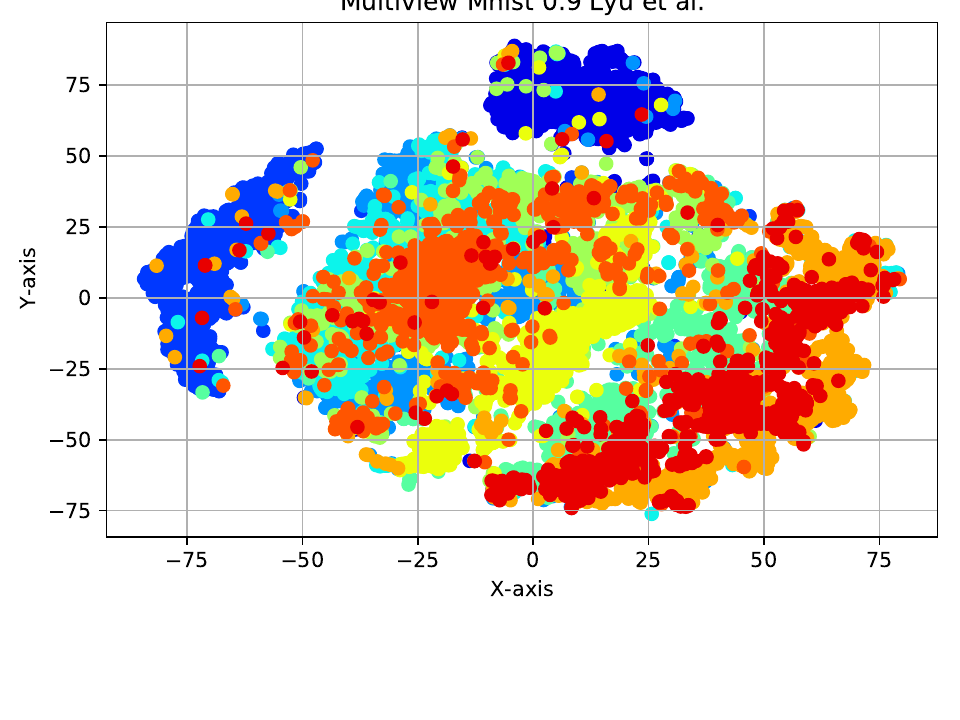}
     \vspace*{-6mm}
    \caption{t-SNE visualizations of the embeddings of the Multiview MNIST dataset for the method of \cite{lyu2021understanding} for $\beta=0$ and different values of $\lambda$ ($\lambda=0.1$-top to $\lambda=0.9$-bottom).}
    \label{fig:LyuMnist0}
    \end{minipage}
\end{figure}

 \clearpage
\begin{figure}
    \centering
    \begin{minipage}{.5\textwidth}
    \includegraphics[scale = 0.39]{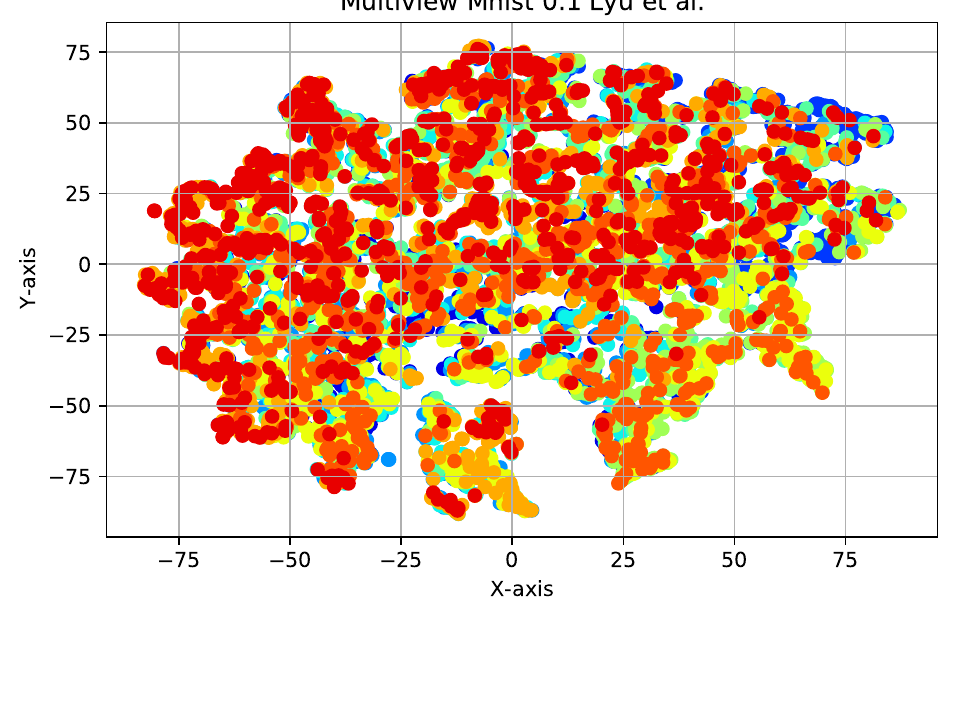}
    \includegraphics[scale = 0.39]{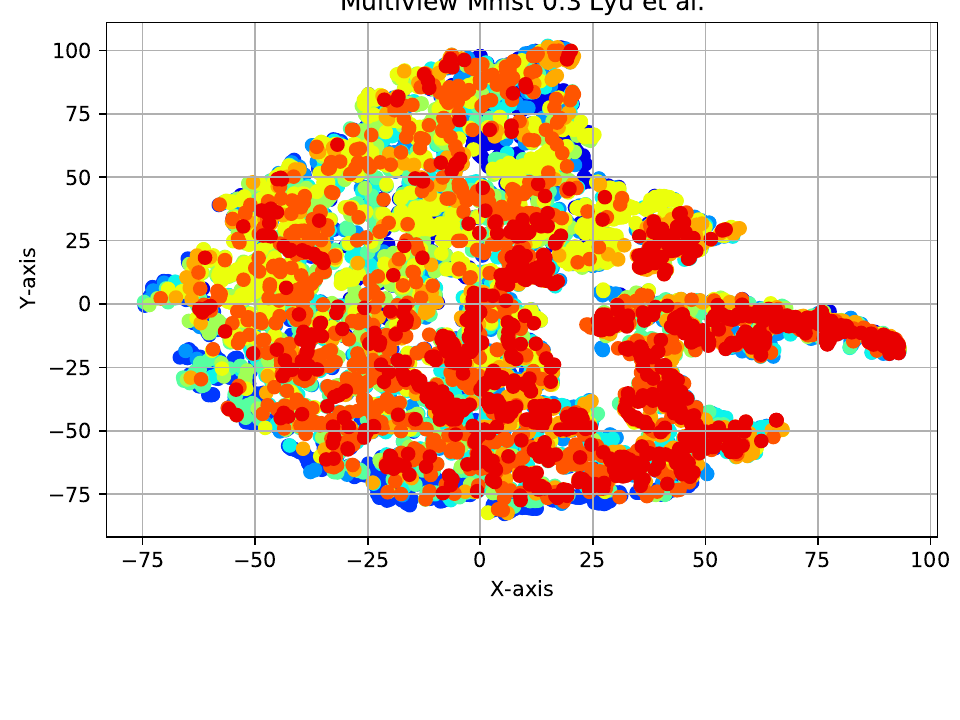}
    \includegraphics[scale = 0.39]{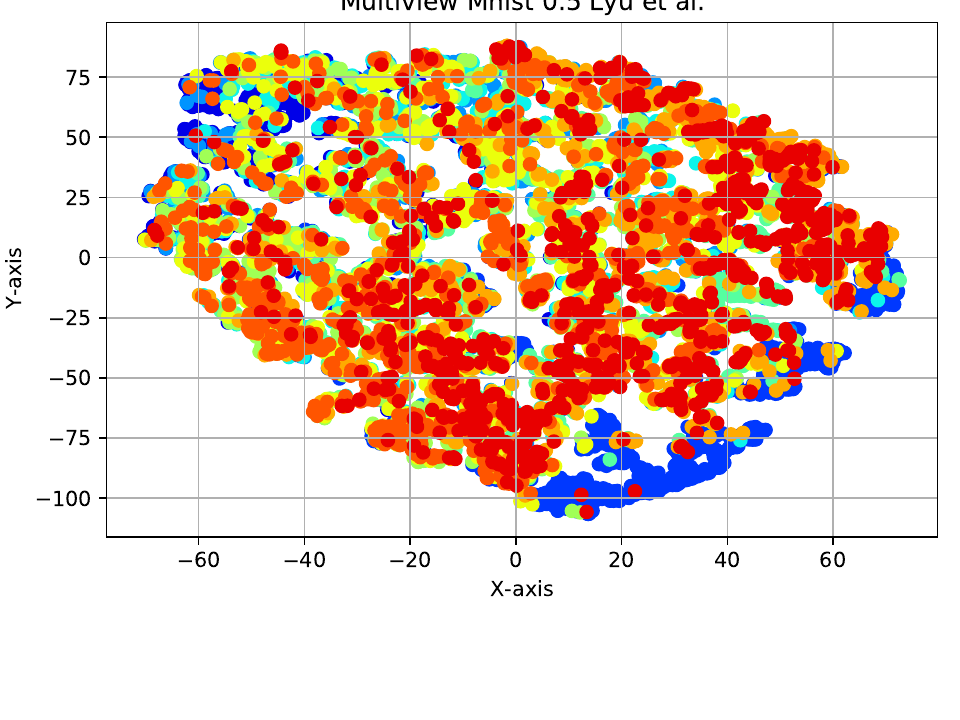}
    \includegraphics[scale = 0.39]{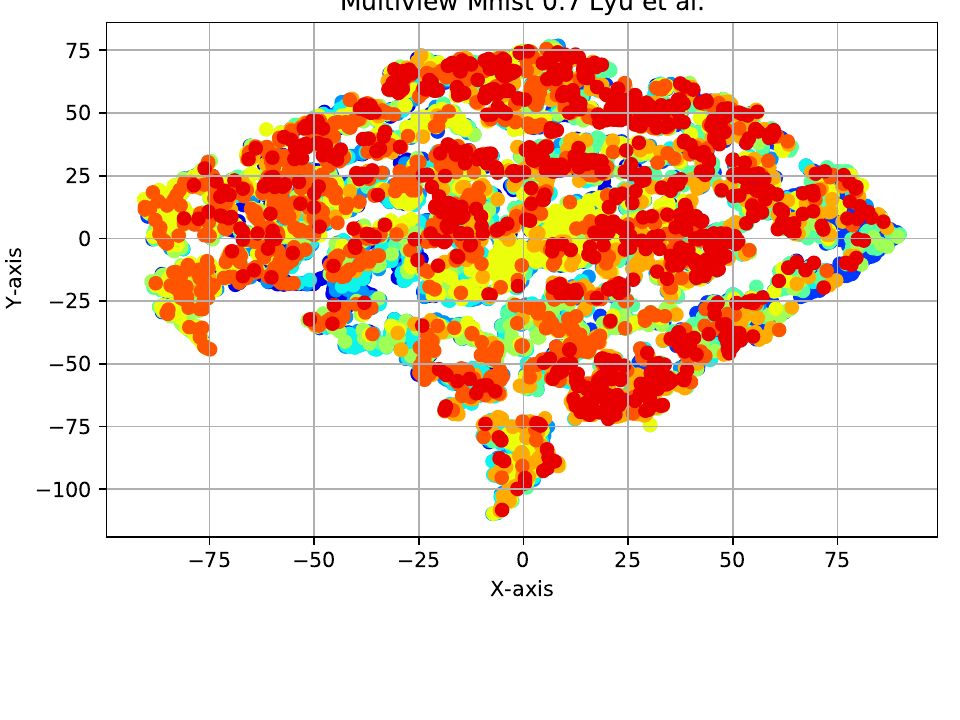}
    \includegraphics[scale = 0.39]{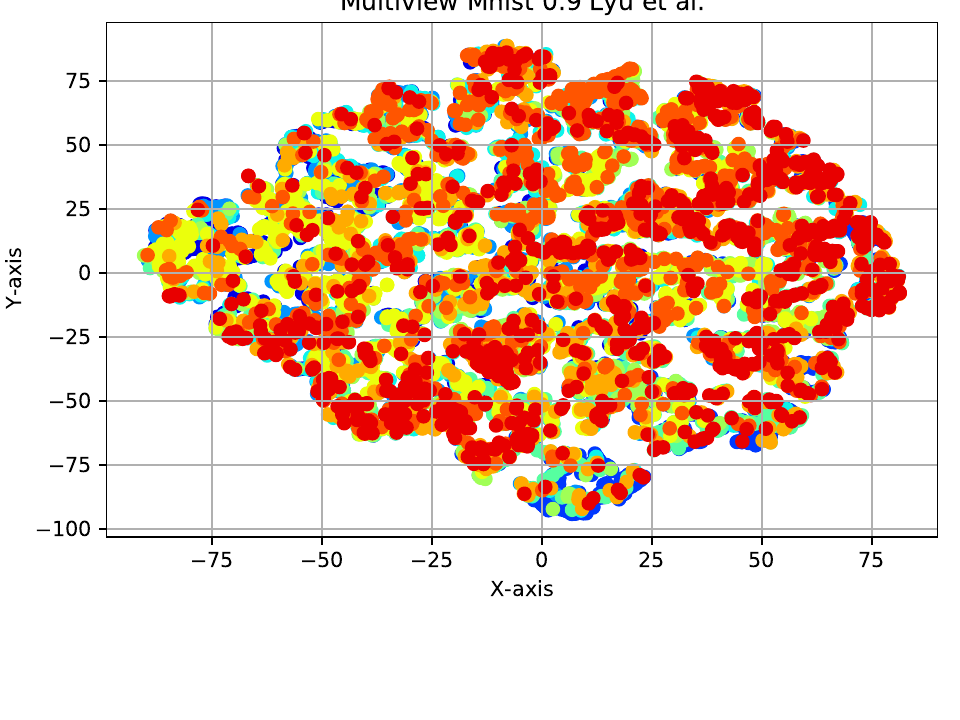}
    \vspace*{-6mm}
    \caption{t-SNE visualizations of the embeddings of the Multiview\\ MNIST dataset for the method of \cite{lyu2021understanding} for $\beta=10^{-3}$ and different\\ values of $\lambda$ ($\lambda=0.1$-top to $\lambda=0.9$-bottom).}
    \label{fig:LyuMnist1}
    \end{minipage}%
    \begin{minipage}{.5\textwidth}
    \centering
    \includegraphics[scale = 0.39]{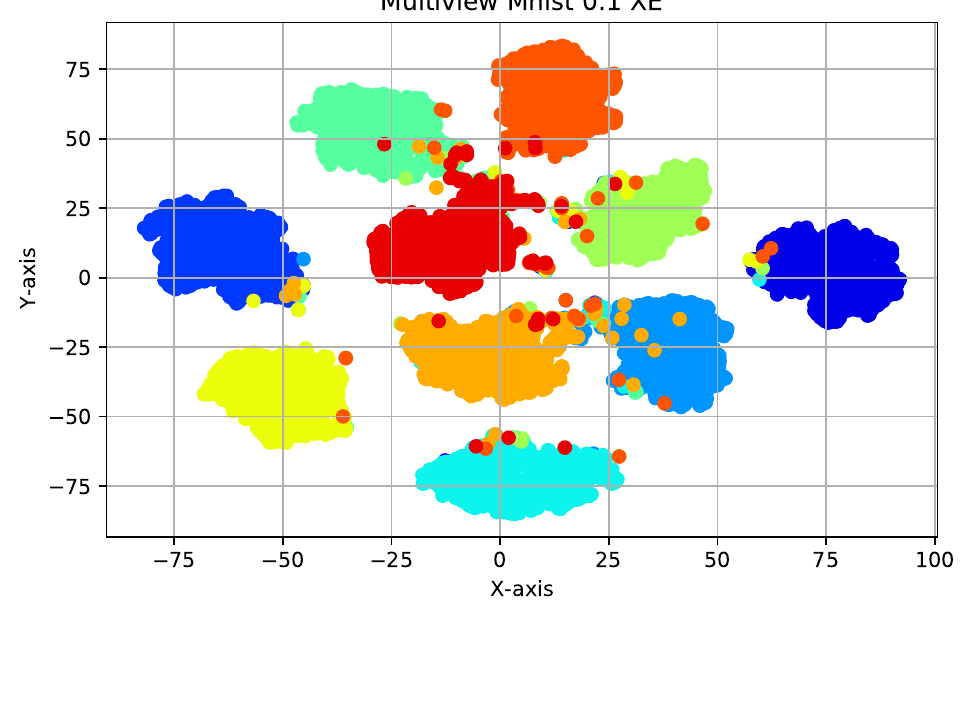}
    \includegraphics[scale = 0.39]{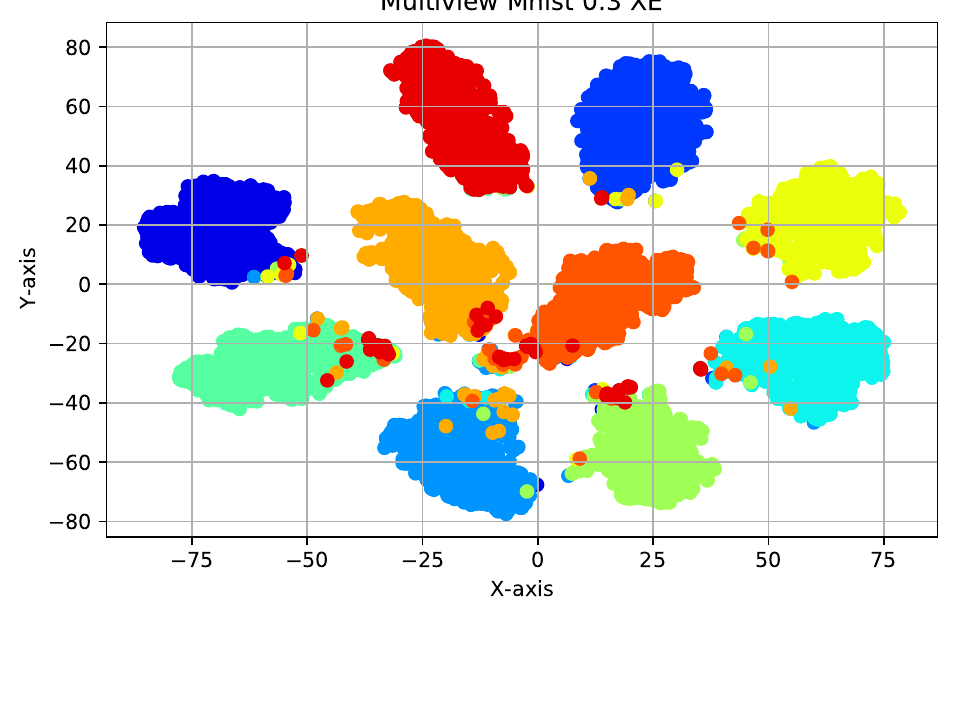}
    \includegraphics[scale = 0.39]{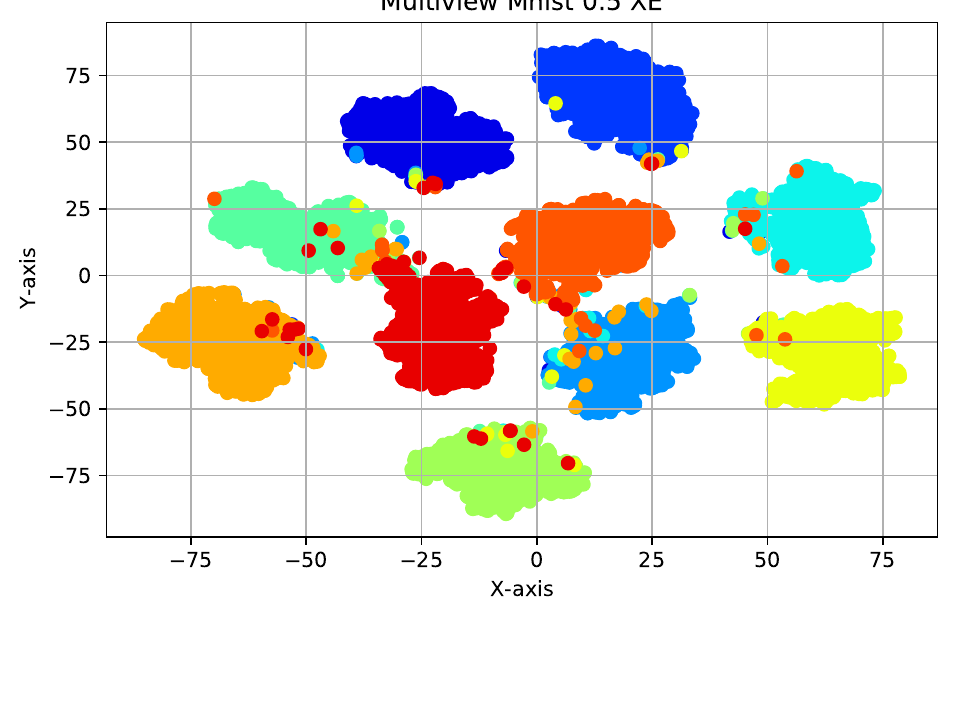}
    \includegraphics[scale = 0.39]{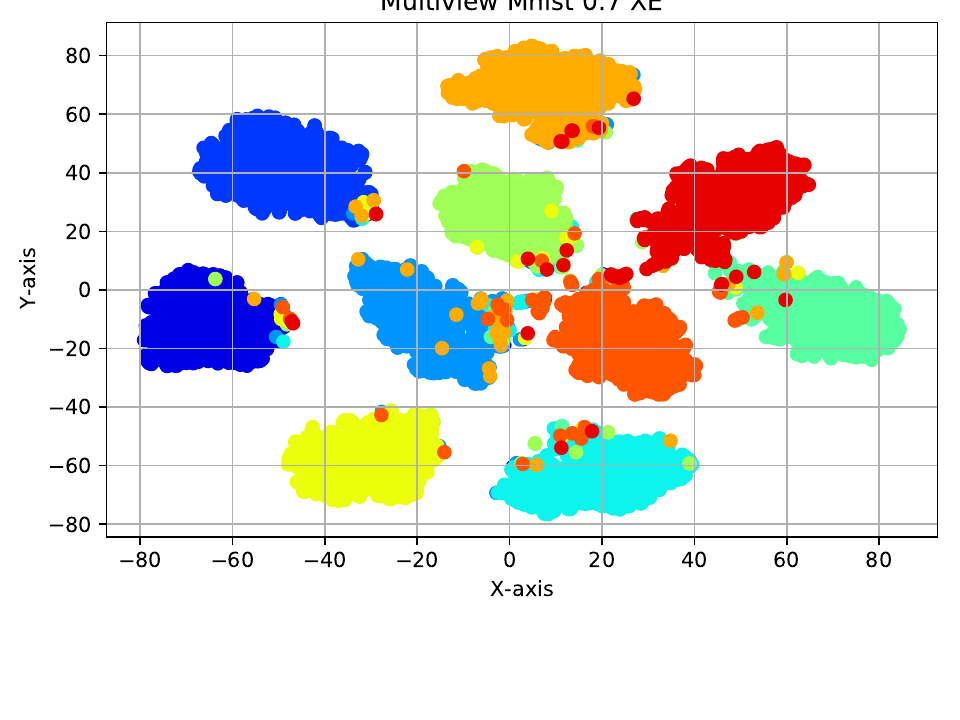}
    \includegraphics[scale = 0.39]{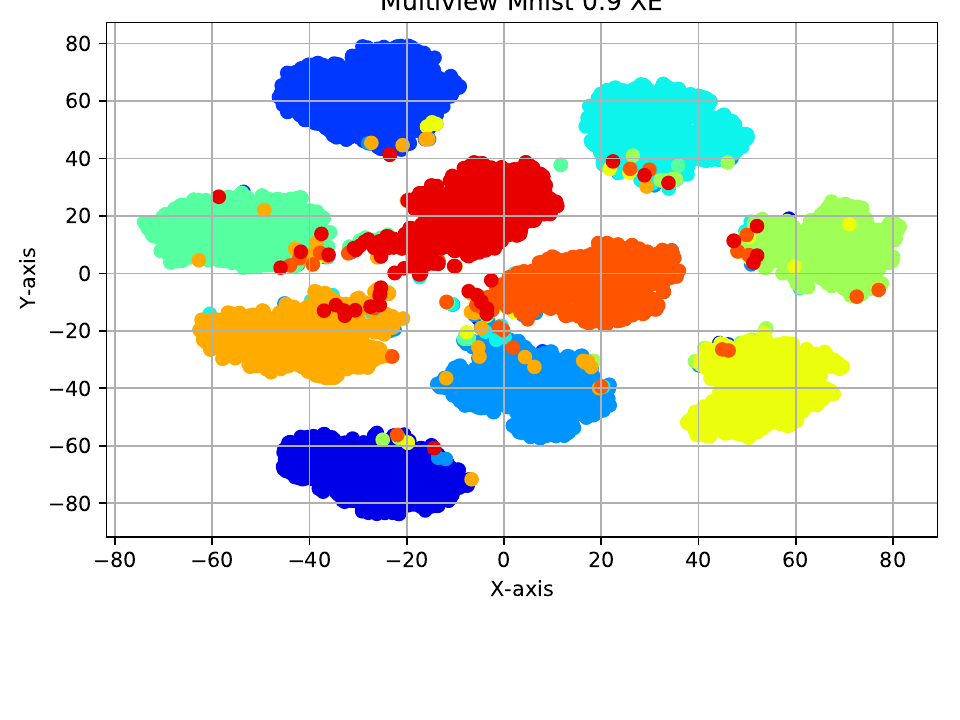}
    \vspace*{-6mm}
    \captionsetup{justification=centering}
    \caption{t-SNE visualizations of the embeddings of the Multiview MNIST dataset for the proposed method for different values of $\lambda$ ($\lambda=0.1$-top to $\lambda=0.9$-bottom).}
    \label{fig:XEMnist}
    \end{minipage}
\end{figure}

\end{document}